%% file: msg.tex
\documentclass[10pt,twocolumn,letterpaper]{article}

\usepackage{cvpr}
\usepackage{times}
\usepackage{epsfig}
\usepackage{graphicx}
\usepackage{subcaption}
\usepackage{multirow}
\usepackage{amsmath}
\usepackage{amssymb}
\usepackage{comment}
\usepackage{enumitem}

\input{includes}

\usepackage[breaklinks=true,bookmarks=false]{hyperref}

\cvprfinalcopy 

\def\cvprPaperID{4679} 

\ifcvprfinal\fi
\begin{document}

\title{MSG-GAN: 
Multi-Scale Gradients for Generative Adversarial Networks}

\author{Animesh Karnewar\\
TomTom\\
{\tt\small animesh.karnewar@tomtom.com}
\and
Oliver Wang \\
Adobe Research\\
{\tt\small owang@adobe.com}
}

\twocolumn[{%
\renewcommand\twocolumn[1][]{#1}%
\maketitle
\thispagestyle{empty}
\begin{center}
    \vspace{-.5cm}
    \centering
    \begin{tabular}{*{2}{c@{\hspace{2px}}}}
    \includegraphics[width=.5\linewidth]{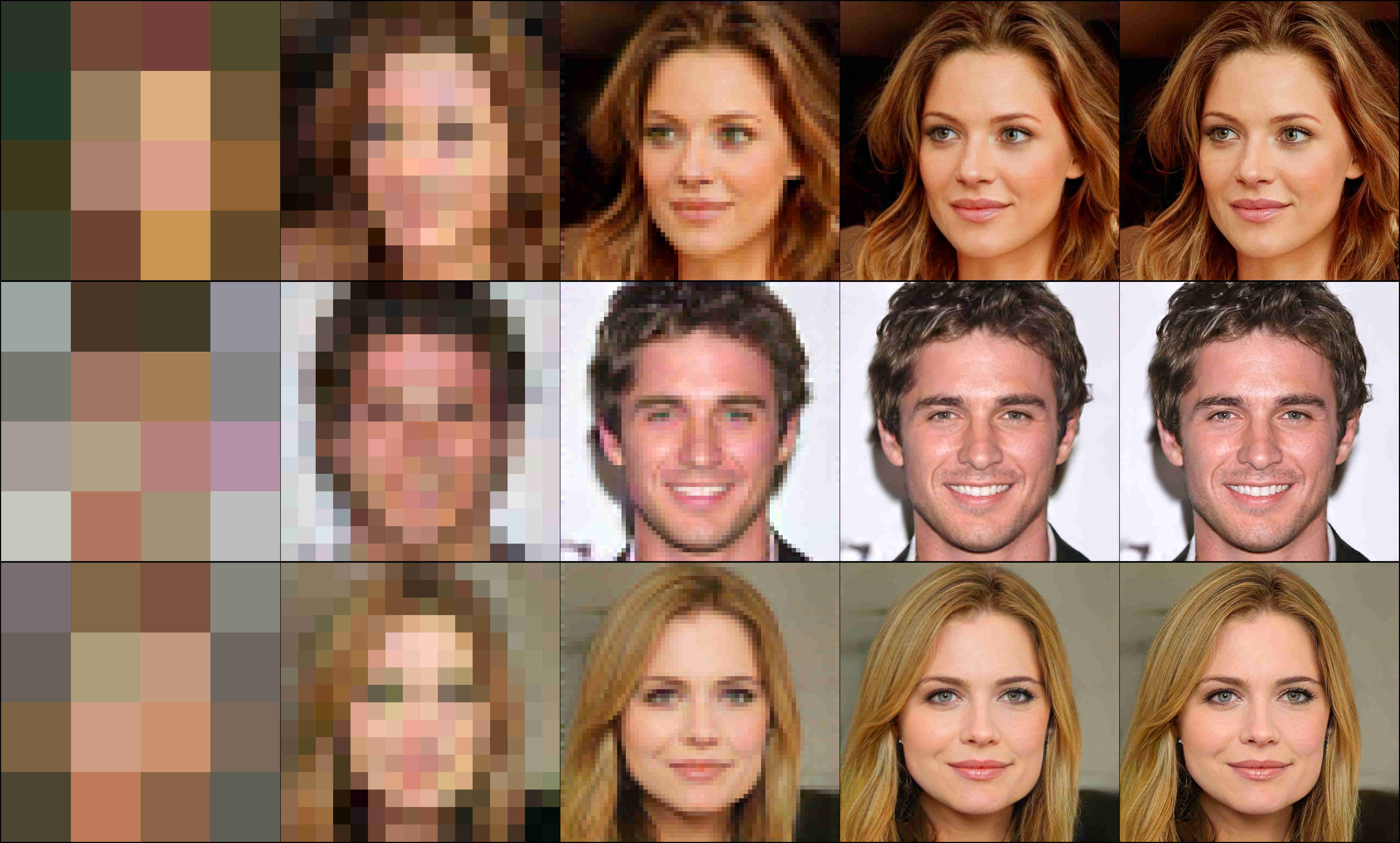} & 
    \includegraphics[width=.5\linewidth]{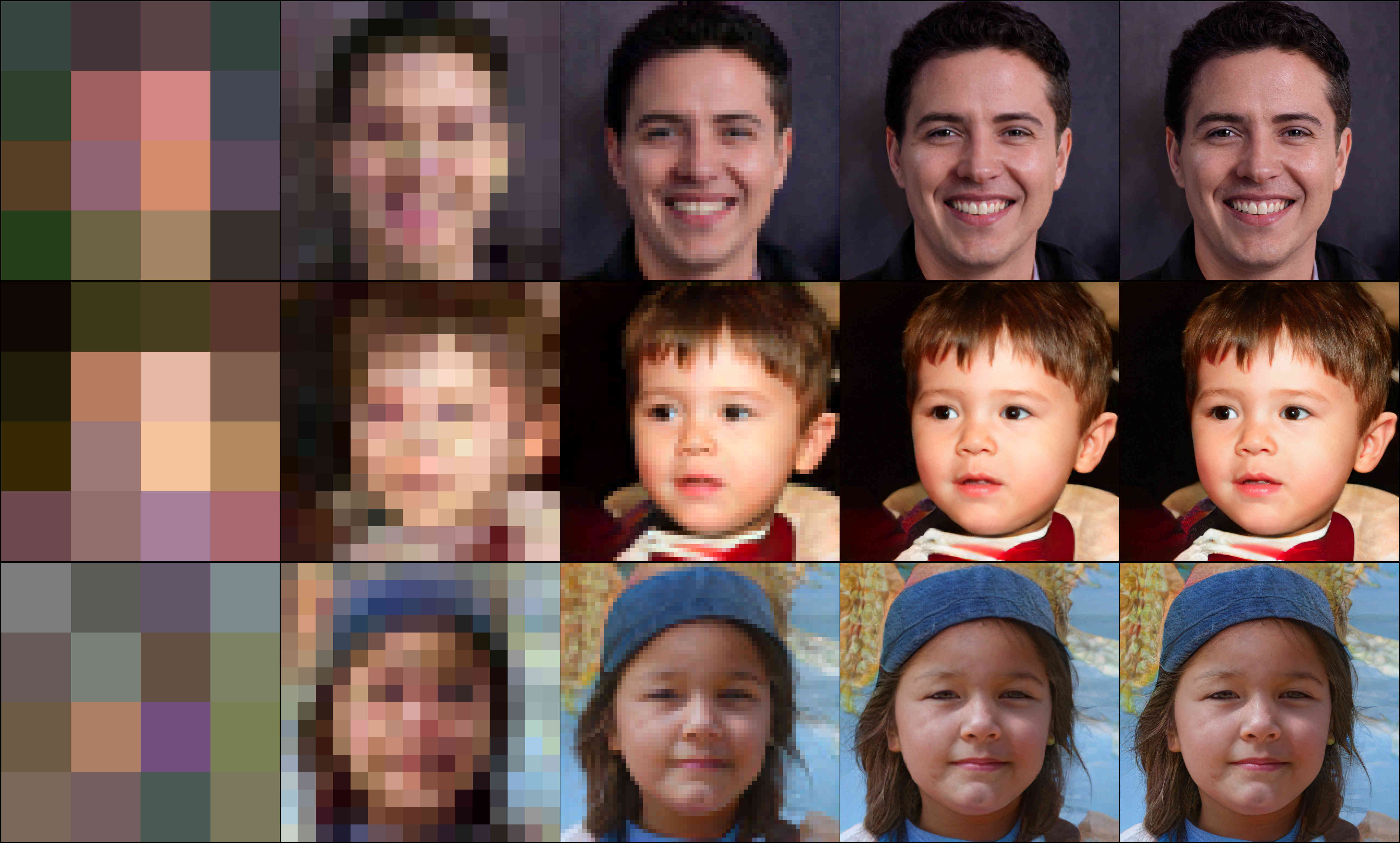} \\
    CelebA-HQ & FFHQ \\
    \end{tabular}
    \captionof{figure}{\label{fig:teaser}Results of our proposed MSG-GAN technique where the generator synthesizes images at all resolutions simultaneously and gradients flow directly to all levels from a single discriminator. 
    The first column has a resolution of \texttt{4x4} which increases towards the right reaching the final output resolution of \texttt{1024x1024}. Best viewed zoomed in on screen.}
\end{center}%
}]

\begin{abstract}
\vspace{-.1cm}
While Generative Adversarial Networks (GANs) have seen huge successes in image synthesis tasks, they are notoriously difficult to adapt to different datasets, in part due to instability during training and sensitivity to hyperparameters. 
One commonly accepted reason for this instability is that gradients passing from the discriminator to the generator become uninformative when there isn't enough overlap in the supports of the real and fake distributions. 
In this work, we propose the Multi-Scale Gradient Generative Adversarial Network (MSG-GAN), a simple but effective technique for addressing this by allowing the flow of gradients from the discriminator to the generator at multiple scales.
This technique provides a stable approach for high resolution image synthesis, and serves as an alternative to the commonly used progressive growing technique.
We show that MSG-GAN converges stably on a variety of image datasets of different sizes, resolutions and domains, as well as different types of loss functions and architectures, all with the same set of \emph{fixed} hyperparameters.
When compared to state-of-the-art GANs, our approach matches or exceeds the performance in most of the cases we tried.
\end{abstract}

\section{Introduction}
\label{sec:introduction}
\input{introduction}

\section{Multi-Scale Gradient GAN}
\label{sec:method}
\input{method}

\section{Experiments}
\label{sec:experimentation}

\input{experiments}

\section{Discussion}
\label{sec:discussion}

\input{discussion}

\section{Acknowledgements}
We would like to thank Alexia Jolicoeur-Martineau (Ph.D. student at MILA) for her guidance over Relativism in GANs and for proofreading the paper. 
Finally we extend special thanks to Michael Hoffman (Sr. Mgr. Software Engineering, TomTom) for his support and motivation.

{\small
\bibliographystyle{ieee_fullname}
\bibliography{msg}
}

\input{msg_supp}

\end{document}

%% file: includes.tex
\usepackage{booktabs} 
\usepackage{colortbl}
\usepackage{tabularx}
\usepackage{multirow}
\usepackage[table,x11names,dvipsnames,table]{xcolor}

\definecolor{rowblue}{RGB}{220,230,240}
\definecolor{myorchid}{RGB}{150,10,30}
\definecolor{myblue}{RGB}{10,30,250}
\definecolor{mygreen}{RGB}{10,120,10}

\DeclareGraphicsExtensions{.jpg,.pdf,.png}

%% file: introduction.tex

\begin{figure*}[t]
\begin{center}
\includegraphics[width=1.0\linewidth]{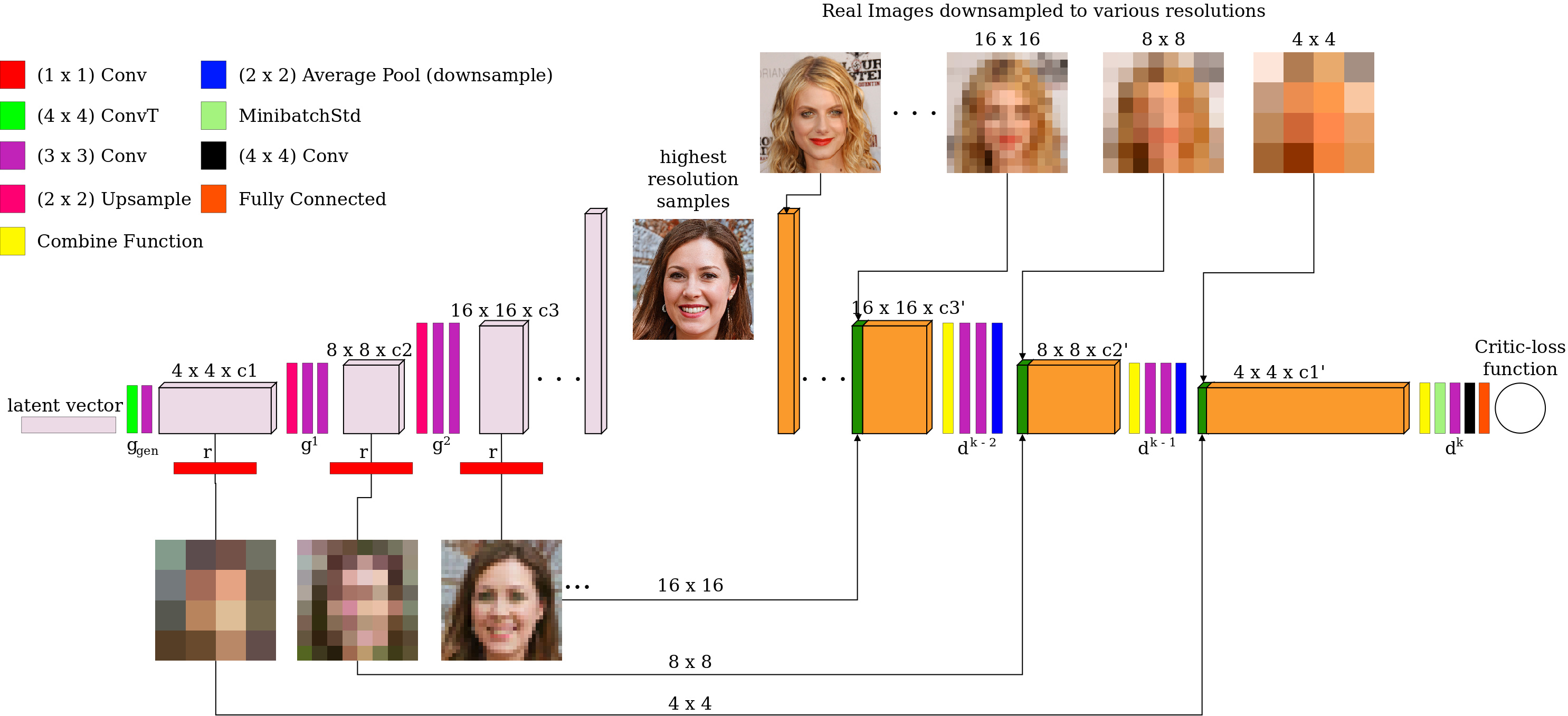}
\end{center}
   \caption{
   Architecture of MSG-GAN, shown here on the base model proposed in ProGANs~\cite{proGAN2018}. 
   Our architecture includes connections from the intermediate layers of the generator to the intermediate layers of the discriminator.
   Multi-scale images sent to the discriminator are concatenated with the corresponding activation volumes obtained from the main path of convolutional layers followed by a combine function (shown in yellow).}
\label{fig:arch}
\end{figure*}

Since their introduction by Goodfellow \etal~\cite{gan2014}, Generative Adversarial Networks (GANs) have become the de facto standard for high quality image synthesis.
The success of GANs comes from the fact that they do not require manually designed loss functions for optimization, and can therefore learn to generate complex data distributions without the need to be able to explicitly define them. 
While flow-based models such as \cite{nice2014, realNVP2016, normalizingFlows2015, glow2018} and autoregressive models such as \cite{pixelRNN2016, pixelCNN2016, pixelCNNpp2017} allow training generative models directly using Maximum Likelihood Estimation (explicitly and implicitly respectively), the fidelity of the generated images has not yet been able to match that of the state-of-the-art GAN models \cite{proGAN2018, styleGAN2018, stylegan2, brock2018large}. 
However, GAN training suffers from two prominent problems: (1) mode collapse and (2) training instability.

The problem of mode collapse occurs when the generator network is only able to capture a subset of the variance present in the data distribution. 
Although numerous works \cite{improvedGAN2016, zhu2017toward, proGAN2018, pacGAN2018} have been proposed to address this problem, it remains an open area of study.
In this work, however, we address the problem of \emph{training instability}.
This is a fundamental issue with GANs, and has been widely reported by previous works~\cite{improvedGAN2016, LSGAN2016, WGAN2017, WGANGP2017, kodali2017convergence, hingeGAN2017, alexiaRelativism2018, proGAN2018, ema2019,  vdb2019}.
We propose a method to address training instability for the task of image generation by investigating how gradients at multiple scales can be used to generate high resolution images (typically more challenging due to the data dimensionality) without relying on previous greedy approaches, such as the progressive growing technique~\cite{proGAN2018, styleGAN2018}. 
MSG-GAN allows the discriminator to look at not only the final output (highest resolution) of the generator, but also at the outputs of the intermediate layers (Fig. ~\ref{fig:arch}).
As a result, the discriminator becomes a function of multiple scale outputs of the generator and importantly, \emph{passes gradients} to all the scales simultaneously (more details in section \ref{sec:motivation} and section \ref{sec:method}). 

Furthermore, our method is robust to different loss functions (we show results on WGAN-GP and Non-saturating GAN loss with 1-sided gradient penalty), datasets (we demonstrate results on a wide range of commonly used datasets and a newly created Indian Celebs dataset), and architectures (we integrate the MSG approach with both ProGANs and StyleGAN base architectures). 
Much like progressive growing \cite{proGAN2018}, we note that multi-scale gradients account for a considerable improvement in FID score over the vanilla DCGAN architecture.
However, our method achieves better performance with comparable training time to state-of-the-art methods on most existing datasets without requiring the extra hyperparameters that progressive growing introduces, such as training schedules and learning rates for different generation stages (resolutions).
This robustness allows the MSG-GAN approach to be easily used ``out-of-the-box'' on new datasets.
We also show the importance of the multi-scale connections on multiple generation stages (coarse, medium, and fine), through ablation experiments on the high resolution FFHQ dataset.

In summary, we present the following contributions. First, we introduce a multiscale gradient technique for image synthesis that improves the stability of training as defined in prior work.
Second, we show that we can robustly generate high quality samples on a number of commonly used datasets, including CIFAR10, Oxford102 flowers, CelebA-HQ, LSUN Churches, Flickr Faces HQ and our new Indian Celebs all with the same \emph{fixed} hyperparameters. This makes our method easy to use in practice.

\subsection{Motivation}
\label{sec:motivation}
Arjovsky and Bottou~\cite{supportOverlapProblem2017} pointed out that one of the reasons for the training instability of GANs is due to the passage of random (uninformative) gradients from the discriminator to the generator when there is insubstantial overlap between the supports of the real and fake distributions. 
Since the inception of GANs, numerous solutions have been proposed to this problem. 
One early example proposes adding instance noise to the real and the fake images so that the supports minimally overlap \cite{supportOverlapProblem2017, Snderby2016AmortisedMI}.
More recently, Peng \etal~\cite{vdb2019} proposed a mutual information bottleneck between input images and the discriminator's deepest representation of those input images called the variational discriminator bottleneck (VDB)~\cite{vdb2019}, and Karras \etal~\cite{proGAN2018} proposed a progressive growing technique to add continually increasing resolution layers. 
The VDB solution forces the discriminator to focus only on the most discerning features of the images for classification, which can be viewed as an adaptive variant of instance noise.
Our work is orthogonal to the VDB technique, and we leave an investigation into a combination of MSG-GAN and VDB to future work. 

The progressive growing technique tackles the instability problem by training the GAN layer-by-layer by gradually doubling the operating resolution of the generated images.
Whenever a new layer is added to the training it is slowly faded in such that the learning of the previous layers are retained.
Intuitively, this technique helps with the support overlap problem because it first achieves a good distribution match on lower resolutions, where the data dimensionality is lower, and then \emph{partially-initializes} (with substantial support overlap between real and fake distributions) higher resolution training with these previously trained weights, focusing on learning finer details.

While this approach is able to generate state-of-the-art results, it can be hard to train, due to the addition of hyperparameters to be tuned per resolution, including different iteration counts, learning rates (which can be different for the Generator and Discriminator \cite{ttur2017}) and the fade-in iterations. In addition, a concurrent submission~\cite{stylegan2} discovered that it leads to phase artifacts where certain generated features are attached to specific spatial locations. 
Hence our main motivation lies in addressing these problems by providing a simpler alternative that leads to high quality results and stable training.

Although the current state-of-the-art in class conditional image generation on the Imagenet dataset, i.e. BigGAN \cite{biggan}, doesn't employ multi-scale image generation, note that the highest resolution they operate on is \texttt{512x512}. 
All high resolution state-of-the-art methods \cite{proGAN2018, styleGAN2018, stylegan2, pix2pixHD, hdgan} use some or the other form of multi-scale image synthesis.   
Multi-scale image generation is a well established technique, with methods existing well before deep networks became popular for this task~\cite{lefebvre2005parallel,wexler2007space}.
More recently, a number of GAN-based methods break the process of high resolution image synthesis into smaller subtasks~\cite{lrgan2017, stackGAN2017, stackGANpp2018, gman2016, madGAN2018, proGAN2018, pix2pixHD, hdgan}.
For example, LR-GAN~\cite{lrgan2017} uses separate generators for synthesizing the background, foreground and compositor masks for the final image.
Works such as GMAN and StackGAN employ a single generator and multiple discriminators for variation in teaching and multi-scale generation respectively~\cite{gman2016, stackGAN2017, stackGANpp2018}.
MAD-GAN~\cite{madGAN2018}, instead uses multiple generators to address mode-collapse by training a multi-agent setup in such a way that different generators capture different modalities in the training dataset. 
LapGAN~\cite{Denton2015DeepGI} models the difference between the generated multi-scale components of a Laplacian pyramid of the images using a single generator and multiple discriminators for different scales. Pix2PixHD \cite{pix2pixHD} uses three architecturally similar discriminators acting upon three different resolutions of the images obtained by downsampling the real and the generated images.

Our proposed method draws architectural inspiration from all these works and builds upon their teachings and ideologies, but has some key differences. 
In MSG-GAN, we use a single discriminator and a single generator with multi-scale connections, which allows for the gradients to flow at multiple resolutions simultaneously.
There are several advantages (driven largely by the simplicity) of the proposed approach. 
If multiple discriminators are used at each resolution~\cite{stackGAN2017, stackGANpp2018, Denton2015DeepGI, hdgan, pix2pixHD}, the total parameters grow exponentially across scales, as repeated downsampling layers are needed, whereas in MSG-GAN the relationship is linear. 
In addition, multiple discriminators with different effective fields \cite{pix2pixHD, hdgan} are not able to share information across scales, which could make the task easier. 
Besides having fewer parameters and design choices required, our approach also avoids the need for an explicit color consistency regularization term across images generated at multiple scales, which was necessary, \eg in StackGAN~\cite{stackGANpp2018}. 

%% file: method.tex

We conduct experiments with the MSG-GAN framework applied to two base architectures, ProGANs~\cite{proGAN2018} and StyleGAN~\cite{styleGAN2018}.
We call these two methods MSG-ProGAN and MSG-StyleGAN respectively.
Despite the name, there is no progressive growing used in any of the MSG variants, and we note that ProGANs without progressive growing is essentially the DCGAN~\cite{radford2015unsupervised} architecture.
Figure~\ref{fig:arch} shows an overview of our MSG-ProGAN architecture, which we define in more detail in this section, and include the MSG-StyleGAN model details in the supplemental material.

Let the initial block of the generator function $g_{gen}$ be defined as $g_{gen}: Z \mapsto A_{begin}$, such that sets $Z$ and $A_{begin}$ are respectively defined as $Z = \mathbb{R}^{512}$, where $ z \sim N(0, \mathbb{I})$ such that $z \in Z$ and $A_{begin} = \mathbb{R}^{4 \times 4 \times 512} $ contains [\texttt{4x4x512}] dimensional activations. 
Let $g^i$ be a generic function which acts as the basic generator block, which in our implementation consists of an upsampling operation followed by two conv layers. 
\begin{align}
& g^{i}: A_{i-1} \mapsto A_{i} \\
& \text{where, } A_{i} = \mathbb{R}^{2^{i + 2} \times 2^{i + 2} \times c_{i}} \\
& \text{and, } i \in \mathbb{N}; A_{0} = A_{\mathit{begin}}
\end{align}
where $c_{i}$ is the number of channels in the $i^{th}$ intermediate activations of the generator. We provide the sizes of $c_{i}$ in all layers in the supplementary material. The full generator $\mathit{GEN}(z)$ then follows the standard format, and can be defined as a sequence of compositions of $k$ such $g$ functions followed by a final composition with $g_{gen}$:
\begin{align}
y' = \mathit{GEN}(z) = g^{k} \circ g^{k - 1} \circ ... g^{i} \circ ... g^{1} \circ g_{gen}(z).
\end{align}

We now define the function $r$ which generates the output at different stages of the generator (red blocks in Fig.~\ref{fig:arch}), where the output corresponds to different downsampled versions of the final output image. 
We model $r$ simply as a (\texttt{1x1}) convolution which converts the intermediate convolutional activation volume into images.
\begin{align}
& r^i : A_{i} \mapsto O_{i}  \\
& \text{where, } O_i = \mathbb{R}^{2^{i + 2} \times 2^{i + 2} \times 3}_{[0 - 1]} \\
& \text{hence, } r^i(g^i(z)) = r^i(a_i) = o_{i}\\
& \text{where, } a_{i} \in A_{i} \text{ and } o_{i} \in O_{i} 
\end{align}
In other words, $o_{i}$ is an image synthesized from the output of the $i^{th}$ intermediate layer of the generator $a_{i}$.
Similar to the idea behind progressive growing~\cite{proGAN2018}, $r$ can be viewed as a regularizer, requiring that the learned feature maps are able to be projected directly into RGB space.

Now we move on to defining the discriminator. 
Because the discriminator's final critic loss is a function of not only the final output of the generator $y'$, but also the intermediate outputs $o_{i}$, gradients can flow from the intermediate layers of the discriminator to the intermediate layers of the generator.
We denote all the components of the discriminator function with the letter $d$. 
We name the final layer of the discriminator (which provides the critic score) $d_{critic}(z')$, and the function which defines the first layer of the discriminator $d^0(y)$ or $d^0(y')$, taking the real image $y$ (true sample) or the highest resolution synthesized image $y'$ (fake sample) as the input. 
Similarly, let $d^j$ represent the intermediate layer function of the discriminator. 
Note that $i$ and $j$ are always related to each other as $j = k - i$.
Thus, the output activation volume $a'_{j}$ of any $j^{th}$ intermediate layer of the discriminator is defined as:
\begin{align}
a'_j &= d^j(\phi\mathit{(o_{k - j}, a'_{j - 1})}) \\
     &= d^j(\phi\mathit{(o_{i}, a'_{j - 1})}),
\end{align}
where $\phi$ is a function used to combine the output $o_{i}$ of the $(i)^{th}$ intermediate layer of the generator (or correspondingly downsampled version of the highest resolution real image $y$) with the corresponding output of the $(j - 1)^{th}$ intermediate layer in the discriminator.
In our experiments, we experimented with three different variants of this combine function:
\begin{align}
\phi_{\mathit{simple}}(x_1, x_2)   &= [x_1; x_2] \label{phi_simple}\\
\phi_{\mathit{lin\_cat}}(x_1, x_2) &= [r'(x_1); x_2] \\
\phi_{\mathit{cat\_lin}}(x_1, x_2) &= r'([x_1; x_2])\label{phi_cat_lin}
\end{align}
where, $r'$ is yet another (\texttt{1x1}) convolution operation similar to $r$ and $[;]$ is a simple channelwise concatenation operation. 
We compare these different combine functions in Sec~\ref{sec:discussion}.

The final discriminator function is then defined as:
\begin{align}
& \mathit{DIS}(y',o_0, o_1, ... o_i, ... o_{k-1}) = \\
& d_{critic} \circ d^{k}(., o_0) \circ d^{k-1}(., o_1) \circ ... d^{j}(., o_i) \circ ... d^0(y')
\end{align}


We experimented with two different loss functions for the $d_{critic}$ function namely, WGAN-GP \cite{WGANGP2017} which was used by ProGAN \cite{proGAN2018} and Non-saturating GAN loss with 1-sided GP \cite{gan2014, mescheder2018training} which was used by StyleGAN \cite{styleGAN2018}. Please note that since the discriminator is now a function of multiple input images generated by the generator, we modified the gradient penalty to be the average of the penalties over each input.

%% file: experiments.tex
\begin{figure*}[t]

\begin{subfigure}[t]{.325\textwidth}
\includegraphics[width=1.0\linewidth]{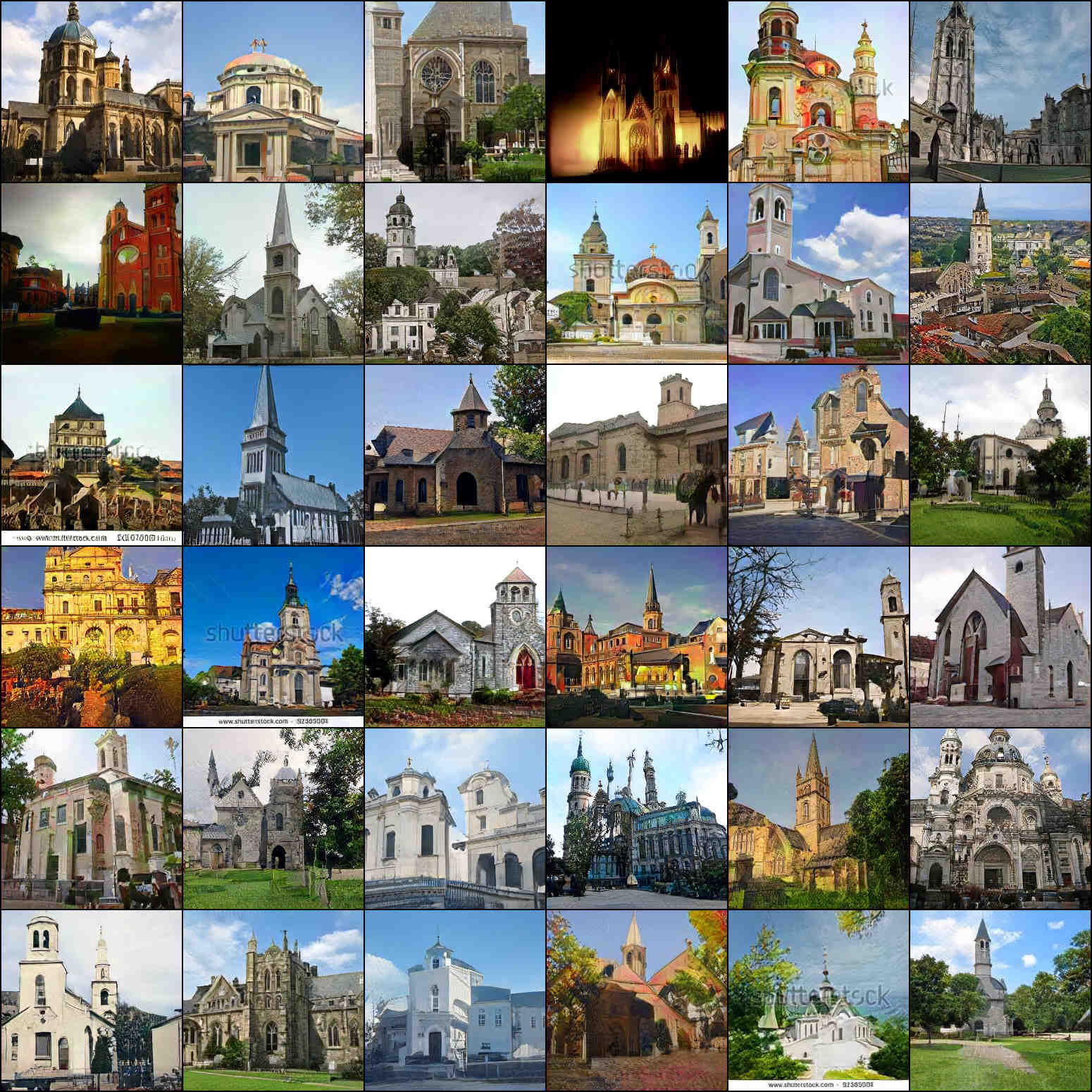}
\caption{LSUN churches}
\end{subfigure}
\hfill
\begin{subfigure}[t]{.325\textwidth}
\centering
\includegraphics[width=1.0\linewidth]{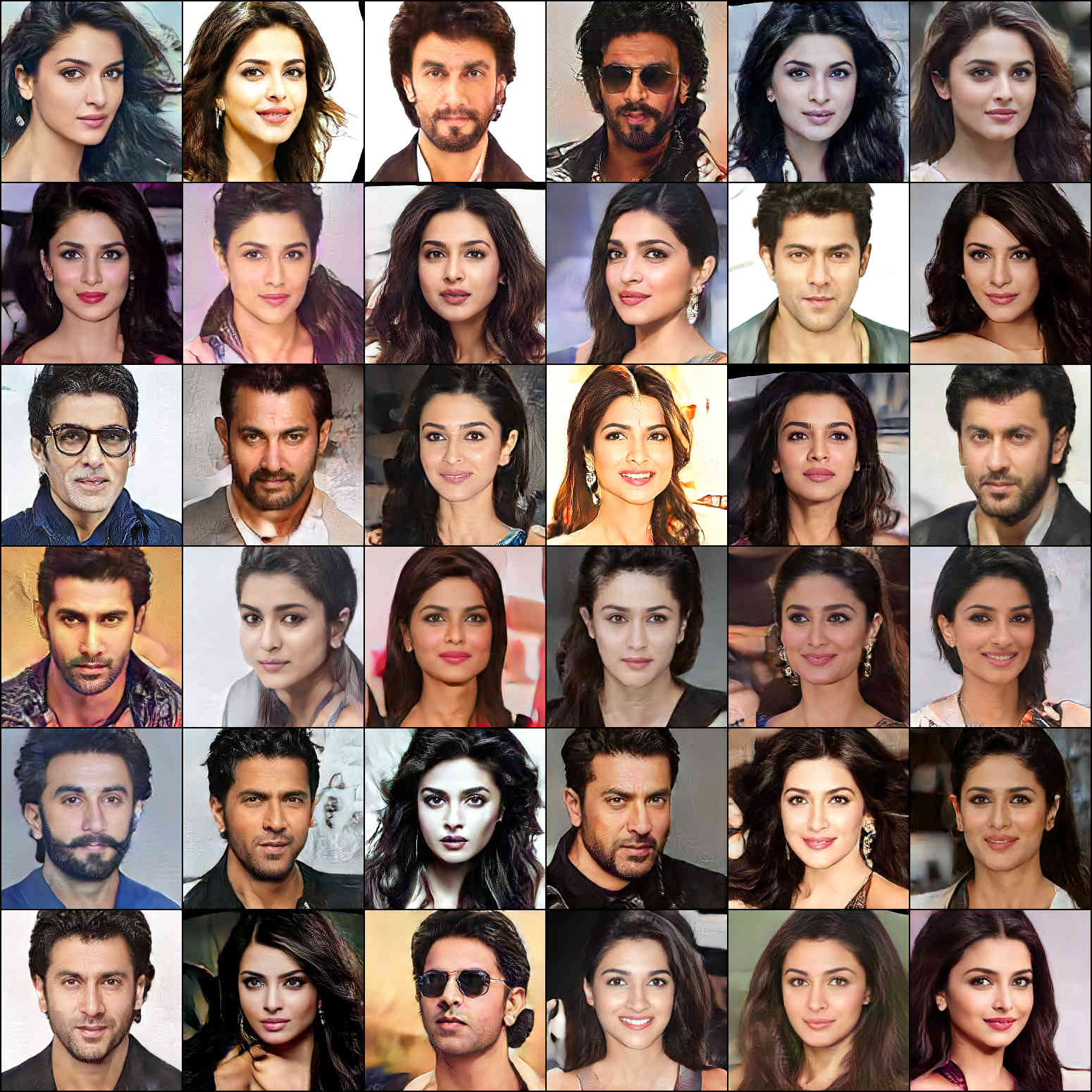}
\caption{Indian Celebs}
\end{subfigure}
\hfill
\begin{subfigure}[t]{.325\textwidth}
\centering
\includegraphics[width=1.0\linewidth]{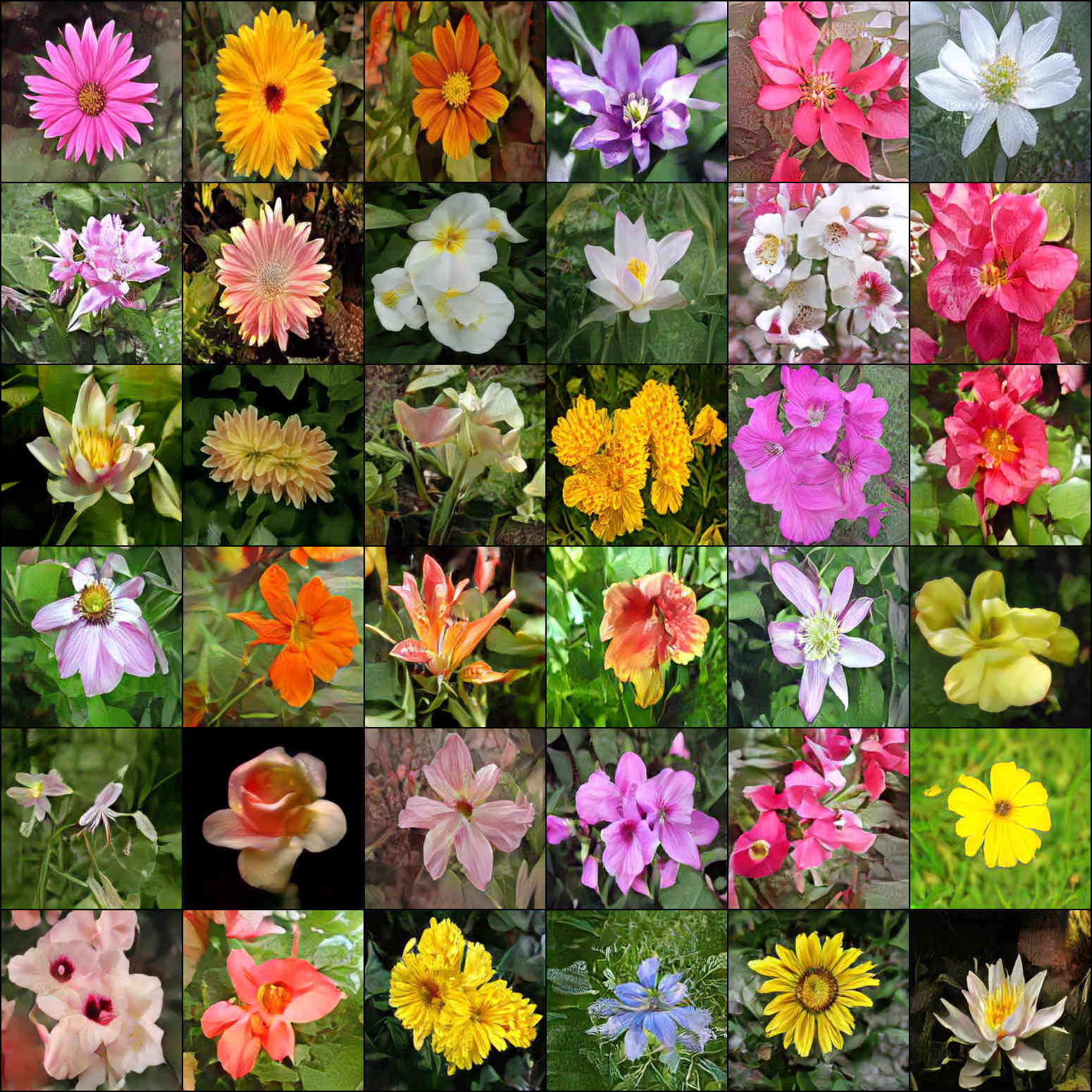}
\caption{Oxford Flowers}
\end{subfigure}

\caption{Random, uncurated samples generated by MSG-StyleGAN on different mid-level resolution (\texttt{256x256}) datasets. Our approach generates high quality results across all datasets with the same hyperparameters. Best viewed zoomed in on screen.}
\label{fig:qualitative_256x256}
\end{figure*}

\begin{table*}[t]
\centering
\resizebox{.9\linewidth}{!}{
\begin{tabular}{lclcccc}
\toprule
Dataset & Size & Method & \# Real Images & GPUs used & Training Time & FID ($\downarrow$) \\
\midrule
Oxford Flowers (\texttt{256x256})   & 8K   & ProGANs$^{*}$   & 10M & 1 V100-32GB  & 104 hrs    & 60.40      \\
                 &      & MSG-ProGAN$^{~}$  & 1.7M   & 1 V100-32GB & 44 hrs & 28.27      \\
                          &      & StyleGAN$^{*}$ & 7.2M & 2 V100-32GB    & 33 hrs & 64.70    \\
                          &      & MSG-StyleGAN & 1.6M & 2 V100-32GB & 16 hrs & \textbf{19.60} \\
\midrule
Indian Celebs (\texttt{256x256})& 3K& ProGANs$^{*}$ & 9M & 2 V100-32GB & 37 hrs & 67.49 \\
                       & & MSG-ProGAN & 2M & 2 V100-32GB & 34 hrs & 36.72 \\
                       & & StyleGAN$^{*}$ & 6M & 4 V100-32GB & 18 hrs  & 61.22 \\
                       &      & MSG-StyleGAN  & 1M & 4 V100-32GB & 7 hrs  &  \textbf{28.44}\\
\midrule
LSUN Churches (\texttt{256x256}) &  126K    & StyleGAN$^{*}$ & 25M & 8 V100-16GB  & 47 hrs & 6.58 \\
                          &      & MSG-StyleGAN & 24M & 8 V100-16GB & 50 hrs & \textbf{5.2}\\
\bottomrule
\end{tabular}
}
\caption{Experiments on mid-level resolution (\ie \texttt{256x256}) datasets. We use author provided scores where possible, and otherwise train models with the official code and recommended hyperparameters (denoted ``$^{*}$'')
}
\label{table:2}
\end{table*}

While evaluating the quality of GAN generated images is not a trivial task, the most commonly used metrics today are the Inception Score (IS, higher is better)~\cite{improvedGAN2016} and Fr\'echet Inception Distance (FID, lower is better)~\cite{ttur2017}.
In order to compare our results with the previous works, we use the IS for the CIFAR10 experiments and the FID for the rest of the experiments, and report the ``number of real images shown'' as done in prior work~\cite{proGAN2018,styleGAN2018}. 

\paragraph{New Indian Celebs Dataset} In addition to existing datasets, we also collect a new dataset consisting of Indian celebrities.
To this end, we collected the images using a process similar to CelebA-HQ.
First, we downloaded images for Indian celebrities by scraping the web for related search queries. 
Then, we detected faces using an off the shelf face-detector and cropped and resized all the images to \texttt{256x256}.
Finally, we manually cleaned the images by filtering out low-quality, erroneous, and low-light images.
In the end, the dataset contained only 3K samples, an order of magnitude less than CelebA-HQ.

\begin{figure*}[t]

\begin{subfigure}[t]{.49\textwidth}
\includegraphics[clip,trim=0 1024 0 0,width=1.0\linewidth]{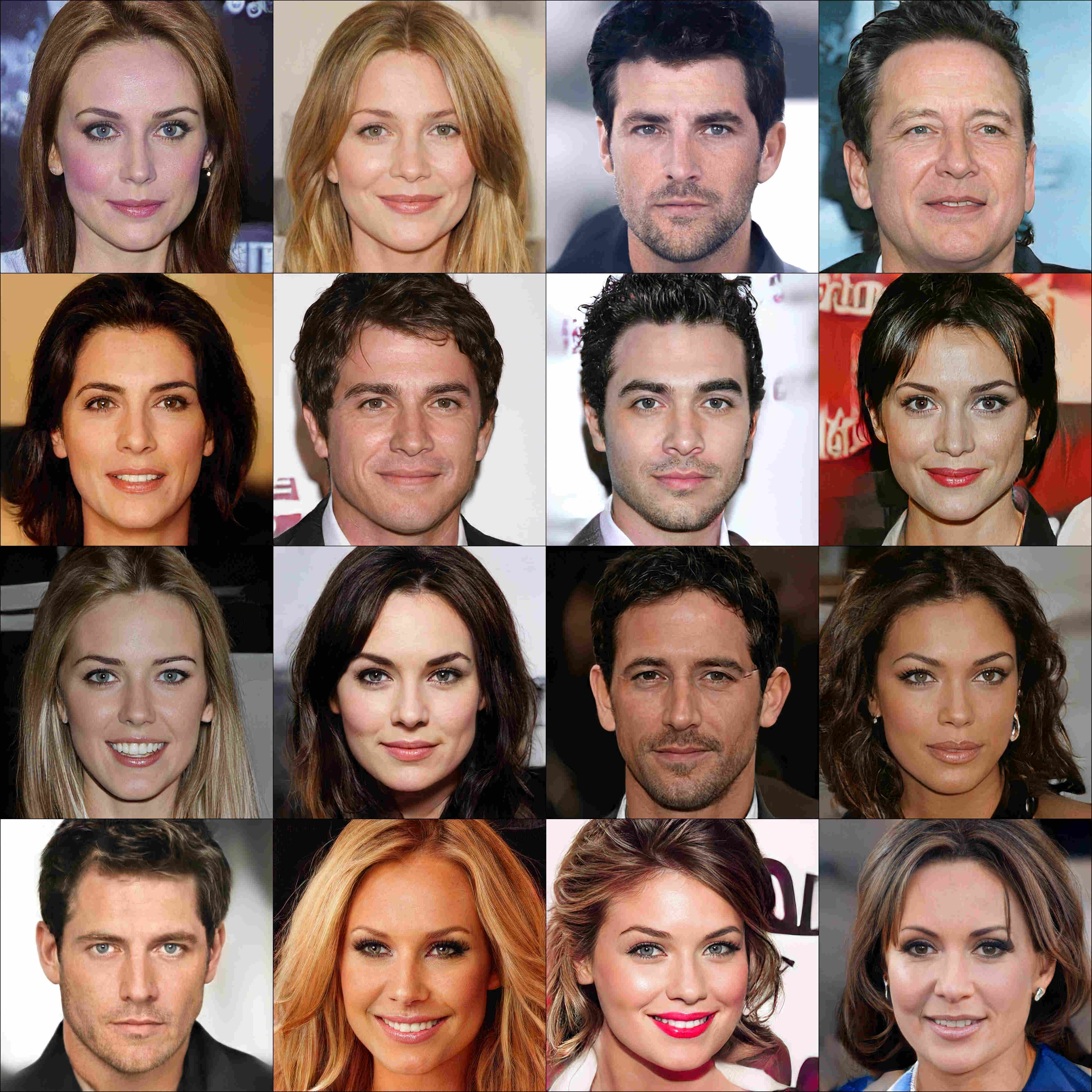}
\caption{CelebA-HQ}
\end{subfigure}
\hfill
\begin{subfigure}[t]{.49\textwidth}
\centering
\includegraphics[clip,trim=0 1024 0 0,width=1.0\linewidth]{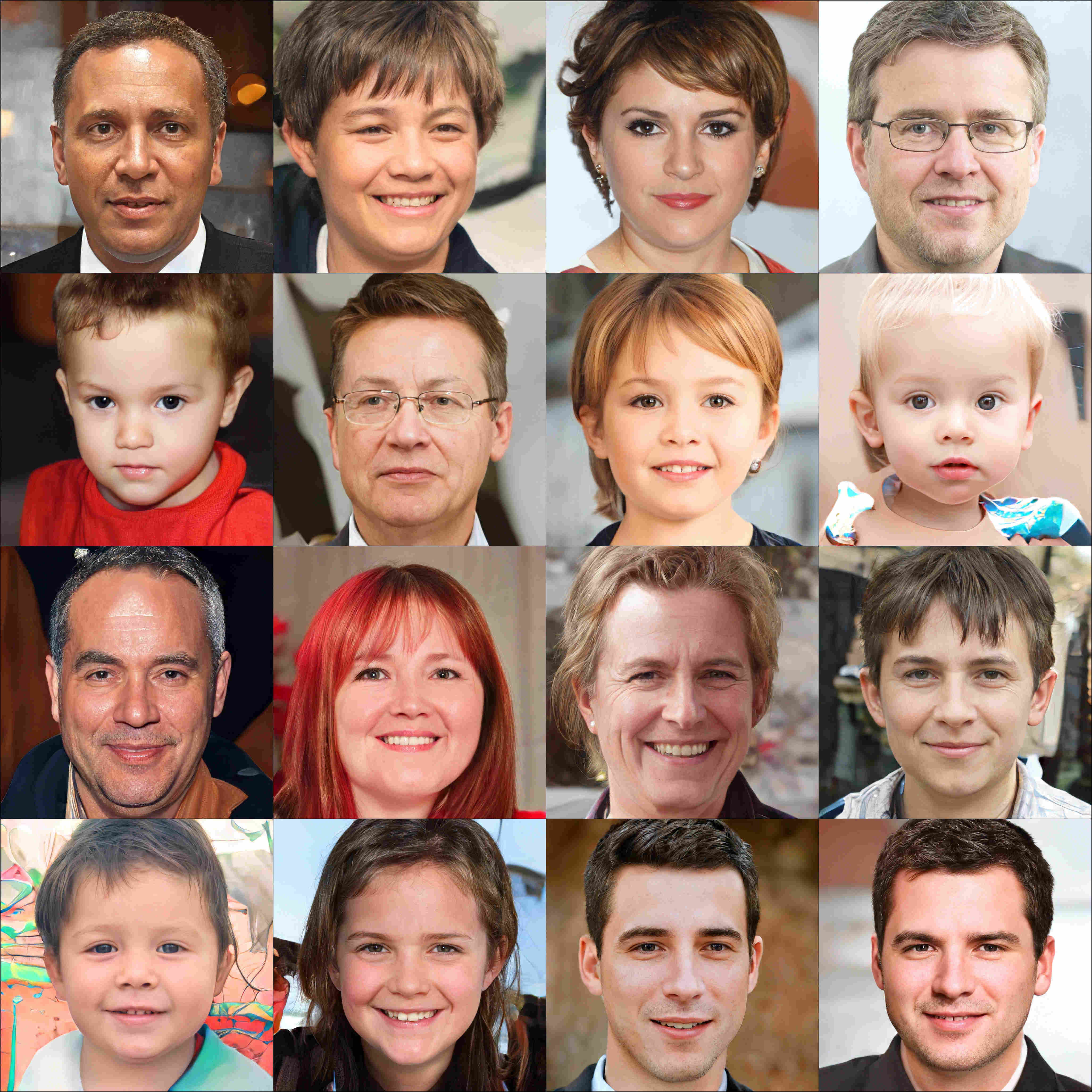}
\caption{FFHQ}
\end{subfigure}

\caption{Random, uncurated samples generated by MSG-StyleGAN on high resolution (\texttt{1024x1024}) datasets. Best viewed zoomed in on screen.}
\label{fig:high_res_samples}
\end{figure*}

\subsection{Implementation Details}
We evaluate our method on a variety of datasets of different resolutions and sizes (number of images); \textbf{CIFAR10} (60K images at \texttt{32x32} resolution); 
\textbf{Oxford flowers} (8K images at \texttt{256x256}), \textbf{LSUN} churches (126K images at \texttt{256x256}), \textbf{Indian Celebs} (3K images at \texttt{256x256} resolution), \textbf{CelebA-HQ} (30K images at \texttt{1024x1024}) and \textbf{FFHQ} (70K images at \texttt{1024x1024} resolution). 

For each dataset, we use the same initial latent dimensionality of \textbf{512}, drawn from a standard normal distribution $ N(0, \mathbb{I}) $ followed by hypersphere normalization \cite{proGAN2018}. 
For all experiments, we use the same hyperparameter settings for MSG-ProGAN and MSG-StyleGAN (lr=0.003), with the only differences being the number of upsampling layers (fewer for lower resolution datasets).

All models were trained with \textbf{RMSprop} (lr=$0.003$) for generator and discriminator. 
We initialize parameters according to the standard normal $ N(0, \mathbb{I}) $ distribution.
To match previously published work, StyleGAN and MSG-StyleGAN models were trained with Non-saturating GAN loss with 1-sided GP while ProGANs and MSG-ProGAN models were trained with the WGAN-GP loss function.

We also extend the MinBatchStdDev technique~\cite{proGAN2018,styleGAN2018}, where the average standard deviation of a batch of activations is fed to the discriminator to improve sample diversity, to our multiscale setup. 
To do this, we add a separate MinBatchStdDev layer at the beginning of each block in the discriminator.
This way, the discriminator obtains batch-statistics of the generated samples along with the straight-path activations at each scale, and can detect some degree of mode collapse by the generator.

When we trained the models ourselves, we report training time and GPUs used. 
We use the same machines for corresponding set of experiments so that direct training time comparisons can be made.
Please note that the variation in numbers of real images shown and training time is because, as is common practice, we report the best FID score obtained in a fixed number of iterations, and the time that it took achieve that score.
All the code and the trained models required for reproducing our work are made available for research purposes at \url{https://github.com/akanimax/msg-stylegan-tf}.

\begin{table*}
\centering
\resizebox{.9\linewidth}{!}{
\begin{tabular}{lclcccc}
\toprule
Dataset & Size & Method & \# Real Images & GPU Used & Training Time & FID ($\downarrow$) \\
\midrule
CelebA-HQ (\texttt{1024x1024})   & 30K & ProGANs \cite{styleGAN2018} & 12M & - & - & \textbf{7.79} \\
            &     & MSG-ProGAN                  & 3.2M  & 8 V100-16GB & 1.5 days & 8.02 \\
\cmidrule{3-7}
         &     & StyleGAN \cite{styleGAN2018} & 25M & - & - &  \textbf{5.17} \\
         &     & MSG-StyleGAN                  & 11M & 8 V100-16GB & 4 days & 6.37 \\
\midrule
FFHQ (\texttt{1024x1024}) & 70K &ProGANs$^{*}$                & 12M  & 4 V100-32GB & 5.5 days  & 9.49      \\
     &     &ProGANs \cite{proGAN2018}    & 12M  & - & -         & \textbf{8.04}      \\
     &     &MSG-ProGAN                   & 6M   & 4 V100-32GB & 6 days    & 8.36      \\
\cmidrule{3-7}
     &     &StyleGAN$^{*}$                & 25M  & 4 V100-32GB & 6 days    & 4.47 \\
     &     &StyleGAN \cite{styleGAN2018}  & 25M  & - &     -     & \textbf{4.40}        \\
     &     &MSG-StyleGAN                  & 9.6M & 4 V100-32GB & 6 days    & 5.8         \\
\bottomrule
\end{tabular}
}
\caption{Experiments on high resolution (\texttt{1024x1024}) datasets. We use author provided scores where possible, and otherwise train models with the official code and recommended hyperparameters (denoted ``$^{*}$'').
}
\label{table:3}
\end{table*}

\subsection{Results}


\paragraph{Quality}\
Table~\ref{table:2} shows quantitative results of our method on various mid-level resolutions datasets. 
Both our MSG-ProGAN and MSG-StyleGAN models achieve better FID scores than the respective baselines of ProGANs and StyleGAN on the (\texttt{256x256}) resolution datasets of Oxford Flowers, LSUN Churches and Indian Celebs.
While each iteration of MSG-GAN is slower than the initial lower resolution iterations of progressive growing, due to all layers being trained together, MSG-GAN tends to converge in fewer iterations, requiring fewer total hours of GPU training time to achieve these scores.
Figure \ref{fig:qualitative_256x256} shows random samples generated on these datasets for qualitative evaluation. 

For high-resolution experiments (Table \ref{table:3}), the MSG-ProGAN model trains in comparable amount of time and gets similar scores on the CelebA-HQ and the FFHQ datasets (\textbf{8.02} vs \textbf{7.79}) and (\textbf{8.36} vs \textbf{8.04}) respectively. 
We note a small difference in the author reported scores and what we were able to achieve with the author provided code. This could be due to subtle hardware differences or variance between runs.
Our MSG-StyleGAN model was unable to beat the FID score of StyleGAN on the CelebA-HQ dataset (\textbf{6.37} vs \textbf{5.17}) and the FFHQ dataset (\textbf{5.8} vs \textbf{4.40}).
We discuss some hypotheses for why this might be in Sec~\ref{sec:discussion}, but note that our method does have other advantages, namely that it seems to be easier to generalize to different datasets as shown in our other experiments.
Also, our generated images do not show any traces of the phase artifacts~\cite{stylegan2} which are prominently visible in progressively grown GANs.

\begin{figure*}
\begin{center}
\includegraphics[width=0.90\linewidth]{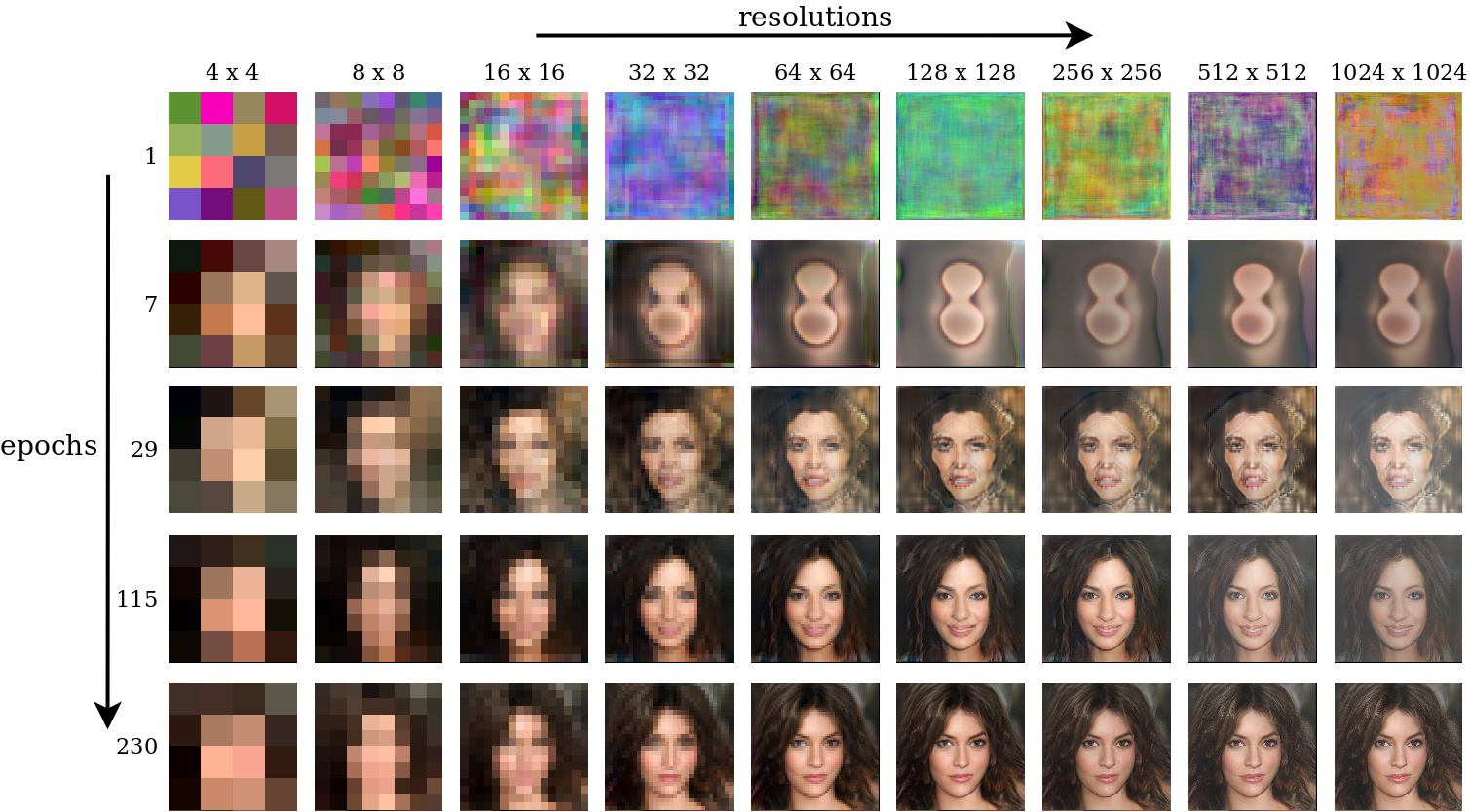}
\end{center}
   \caption{During training, all the layers in the MSG-GAN synchronize across the generated resolutions fairly early in the training and subsequently improve the quality of the generated images at all scales simultaneously. Throughout the training the generator makes only minimal incremental improvements to the images generated from fixed latent points.}
\label{fig:trainingvisualization}
\end{figure*}
\begin{figure*}
\begin{center}
\includegraphics[width=0.48\linewidth]{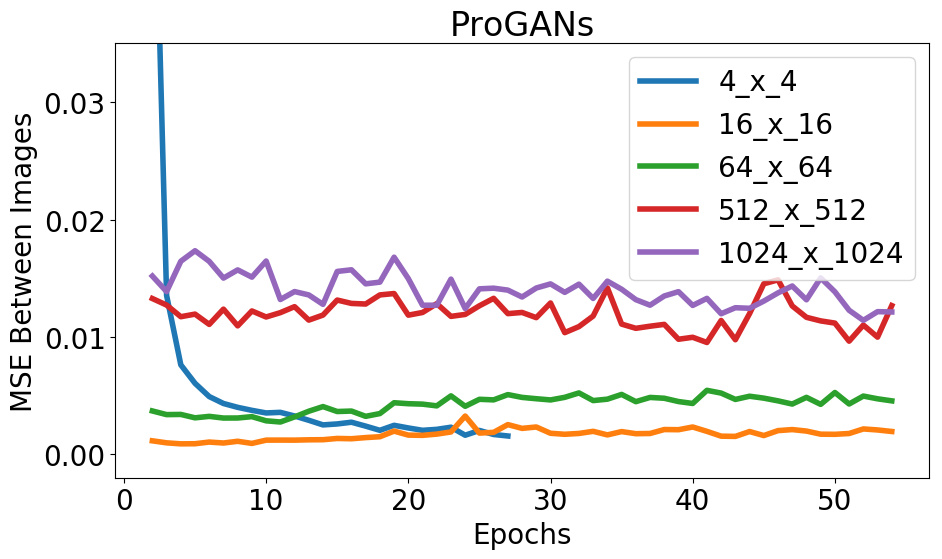}
\includegraphics[width=0.48\linewidth]{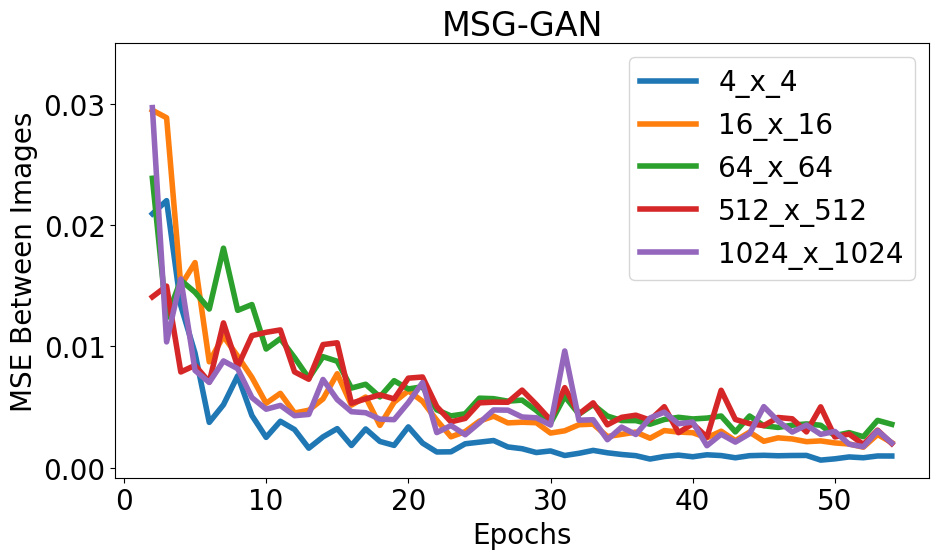}
\end{center}
   \caption{ Image stability during training. These plots show the MSE between images generated from the same latent code at the beginning of sequential epochs (averaged over 36 latent samples) on the CelebA-HQ dataset.
   MSG-ProGAN converges stably over time while ProGANs~\cite{proGAN2018} continues to vary significantly across epochs. 
   }
\label{fig:stability}
\end{figure*}

\paragraph{Stability during training} 
To compare the stability of MSG-ProGAN with ProGANs during training, we measure the changes in the generated samples for the same fixed latent points as iterations progress (on CelebA-HQ dataset).
This method was introduced by \cite{ema2019} as a way to measure stability during training, which we quantify by calculating the mean squared error between two consecutive samples.
Figure \ref{fig:stability} shows that while ProGANs tends towards convergence (making less changes) for lower resolutions only, MSG-ProGAN shows the same convergence trait for all the resolutions. 
The training epochs for the ProGANs take place in sequence over each resolution, whereas for the MSG-ProGAN they are simultaneous (Fig. \ref{fig:trainingvisualization}).
While not necessary for generating good results, methods with high stability can be advantageous in that it is easier to get a reasonable estimate for how the final result will look by visualizing snapshots during training, which can help when training jobs take on the order of days to weeks. 

\begin{table}[t]
\centering
\resizebox{\linewidth}{!}{
\begin{tabular}{llccl}
\toprule
Method & \# Real Images & Learning rate & IS ($\uparrow$) \\
\midrule
Real Images                         & - & - & 11.34 \\
MSG-ProGAN                      	& 12M & 0.003 & \textbf{8.63} \\
MSG-ProGAN                        	& 12M & 0.001 & 8.24 \\
MSG-ProGAN                          & 12M & 0.005 & 8.33 \\
MSG-ProGAN                          & 12M & 0.01  & 7.92 \\
\bottomrule
\end{tabular}
}
\caption{Robustness to learning rate on CIFAR-10. We see that our approach converges to similar IS scores over a range of learning rates. 
}
\label{table:1}
\end{table}

\paragraph{Robustness to learning rate}
\label{sec:ablations}
It has been observed by prior work~\cite{improvedGAN2016, alexiaRelativism2018, metz2016unrolled, mescheder2018training} and also our experience, that convergence of GANs during training is very heavily dependant on the choice of hyperparameters, in particular, learning rate.
To validate the robustness of MSG-ProGAN, we trained our network with four different learning rates (0.001, 0.003, 0.005 and 0.01) for the CIFAR-10 dataset (Table.~\ref{table:1}). 
We can see that all of our four models converge, producing sensible images and similar inception scores, even with large changes in learning rate. 
Robust training schemes are significant as they indicate how easily a method can be generalized to unseen datasets.

%% file: discussion.tex
\begin{table}[t]
\centering
\begin{tabular}{lc}
\toprule
Level of Multi-scale connections & FID ($\downarrow$)\\
\midrule
No connections (DC-GAN)     & 14.20 \\
Coarse Only                 & 10.84 \\
Middle Only                 &  9.17 \\
Fine Only                   &  9.74 \\
All (MSG-ProGAN)            &  \textbf{8.36} \\
ProGAN$^*$                  &  9.49 \\
\bottomrule
\end{tabular}
\caption{Ablation experiments for varying degrees of multiscale gradient connections on the high resolution (\texttt{1024x1024}) FFHQ dataset. Coarse contains connections at (\texttt{4x4}) and (\texttt{8x8}), middle at (\texttt{16x16}) and (\texttt{32x32}); and fine at (\texttt{64x64}) till (\texttt{1024x1024}).}
\label{table:4}
\end{table}

\begin{table}[t]
\centering
\begin{tabular}{llc}
\toprule
Method & Combine function & FID ($\downarrow$)\\
\midrule
MSG-ProGAN & $\phi_{lin\_cat}$ & 11.88     \\
           & $\phi_{cat\_lin}$ & 9.63      \\
           & $\phi_{simple}$   & \textbf{8.36}      \\
\midrule
MSG-StyleGAN & $\phi_{simple}$ & 6.46       \\
             & $\phi_{lin\_cat}$ & 6.12     \\
             & $\phi_{cat\_lin}$ & \textbf{5.80} \\
\bottomrule
\end{tabular}
\caption{Experiments with different combine functions on the high resolution (\texttt{1024x1024}) FFHQ dataset.}
\label{table:5}
\end{table}

\paragraph{Ablation Studies}
We performed two types of ablations on the MSG-ProGAN architecture. 
Table~\ref{table:4} summarizes our experiments on applying ablated versions of the Multi-Scale Gradients, where we only add subsets of the connections from the generator to the discriminator at different scales.
We can see that adding multi-scale gradients at any level to the ProGANs/DCGAN architecture improves the FID score. 
Interestingly, adding only mid-level connections performs slightly better than adding only coarse or fine-level connections, however the overall best performance is achieved with the connections at all levels.

Table~\ref{table:5} presents our experiments with the different variants of the combine function $\phi$ on the MSG-ProGAN and the MSG-StyleGAN architectures. 
$\phi_{simple}$ (Eq~\ref{phi_simple}) performed best on the MSG-ProGAN architecture while the $\phi_{cat\_lin}$ (Eq~\ref{phi_cat_lin}) has the best FID score on the MSG-StyleGAN architecture.
All results shown in this work employ these respective combine functions. 
We can see through these experiments that the combine function also plays an important role in the generative performance of the model, and it is possible that a more advanced combine function such as multi-layer densenet or AdaIN \cite{adain} could improve the results even further.

\paragraph{Limitations and Future Work} 
Our method is not without limitations. 
We note that using progressive training, the first set of iterations at lower resolutions take place much faster, whereas each iteration of MSG-GAN takes the same amount of time. 
However, we observe that MSG-GAN requires fewer total iterations to reach the same FID, and often does so after a similar length of total training time.



In addition, because of our multi-scale modification in MSG-StyleGAN, our approach cannot take advantage of the mixing regularization trick~\cite{styleGAN2018}, where multiple latent vectors are mixed and the resulting image is forced to be realistic by the discriminator. 
This is done to allow the mixing of different styles at different levels at test time, but also improves overall quality. 
Interestingly, even though we do not explicitly enforce mixing regularization, our method is still able to generate plausible mixing results (see supplementary material).
\paragraph{Conclusion}
Although huge strides have been made towards photo-realistic high resolution image synthesis~\cite{brock2018large,styleGAN2018, stylegan2}, true photo-realism has yet to be achieved, especially with regards to domains with substantial variance in appearance. 
In this work, we presented the MSG-GAN technique which contributes to these efforts with a simple approach to enable high resolution multi-scale image generation with GANs.

%% file: msg_supp.tex
\section{Appendix}
\begin{table}[t]
\begin{center}
\footnotesize
\begin{tabular}{|c|l|c|c|}

\hline
Block   & Operation       & Act.    & Output Shape \\
\hline

\multirow{3}{*}{1.} & Latent Vector     & Norm     & 512 x 1 x 1  \\
& Conv 4 x 4    & LReLU     & 512 x 4 x 4   \\
& Conv 3 x 3    & LReLU     & 512 x 4 x 4   \\
\hline

\multirow{3}{*}{2.} & Upsample     & -     & 512 x 8 x 8  \\
& Conv 3 x 3    & LReLU     & 512 x 8 x 8   \\
& Conv 3 x 3    & LReLU     & 512 x 8 x 8   \\
\hline

\multirow{3}{*}{3.} & Upsample     & -     & 512 x 16 x 16  \\
& Conv 3 x 3    & LReLU     & 512 x 16 x 16   \\
& Conv 3 x 3    & LReLU     & 512 x 16 x 16   \\
\hline

\multirow{3}{*}{4.} & Upsample     & -     & 512 x 32 x 32  \\
& Conv 3 x 3    & LReLU     & 512 x 32 x 32   \\
& Conv 3 x 3    & LReLU     & 512 x 32 x 32   \\
\hline

\multicolumn{4}{|c|}{Model 1 $\uparrow$} \\
\hline

\multirow{3}{*}{5.} & Upsample     & -     & 512 x 64 x 64  \\
& Conv 3 x 3    & LReLU     & 256 x 64 x 64   \\
& Conv 3 x 3    & LReLU     & 256 x 64 x 64   \\
\hline

\multirow{3}{*}{6.} & Upsample     & -     & 256 x 128 x 128  \\
& Conv 3 x 3    & LReLU     & 128 x 128 x 128   \\
& Conv 3 x 3    & LReLU     & 128 x 128 x 128   \\
\hline 

\multicolumn{4}{|c|}{Model 2 $\uparrow$} \\
\hline

\multirow{3}{*}{7.} & Upsample     & -     & 128 x 256 x 256  \\
& Conv 3 x 3    & LReLU     & 64 x 256 x 256   \\
& Conv 3 x 3    & LReLU     & 64 x 256 x 256   \\
\hline 

\multicolumn{4}{|c|}{Model 3 $\uparrow$} \\
\hline

\multirow{3}{*}{8.} & Upsample     & -     & 64 x 512 x 512  \\
& Conv 3 x 3    & LReLU     & 32 x 512 x 512   \\
& Conv 3 x 3    & LReLU     & 32 x 512 x 512   \\
\hline 

\multirow{3}{*}{9.} & Upsample     & -     & 32 x 1024 x 1024  \\
& Conv 3 x 3    & LReLU     & 16 x 1024 x 1024   \\
& Conv 3 x 3    & LReLU     & 16 x 1024 x 1024   \\
\hline 

\multicolumn{4}{|c|}{Model full $\uparrow$} \\
\hline

\end{tabular}
\end{center}
\caption{Generator architecture for the MSG-ProGAN models used in training.}
\label{table:gen}
\end{table}

\begin{table}[t]
\begin{center}
\footnotesize
\begin{tabular}{|c|l|c|c|}

\hline
Block   & Operation       & Act.    & Output Shape \\
\hline

\multicolumn{4}{|c|}{Model full $\downarrow$} \\
\hline

   & Raw RGB images 0& -  &  3 x 1024 x 1024 \\
   & FromRGB 0  & -     & 16 x 1024 x 1024 \\
1. & MinBatchStd & -     & 17 x 1024 x 1024 \\
   & Conv 3 x 3  & LReLU & 16 x 1024 x 1024 \\
   & Conv 3 x 3  & LReLU & 32 x 1024 x 1024 \\
   & AvgPool     & -     & 32 x 512 x 512   \\
\hline

   & Raw RGB images 1& -  &  3 x 512 x 512 \\
   & Concat/$\phi_{\mathit{simple}}$ & -     & 35 x 512 x 512 \\
2. & MinBatchStd & -     & 36 x 512 x 512 \\
   & Conv 3 x 3  & LReLU & 32 x 512 x 512 \\
   & Conv 3 x 3  & LReLU & 64 x 512 x 512 \\
   & AvgPool     & -     & 64 x 256 x 256   \\
\hline

\multicolumn{4}{|c|}{Model 3 $\downarrow$} \\
\hline

   & Raw RGB images 2& -  &  3 x 256 x 256 \\
   & Concat/$\phi_{\mathit{simple}}$ & -     & 67 x 256 x 256 \\
3. & MinBatchStd & -     & 68 x 256 x 256 \\
   & Conv 3 x 3  & LReLU & 64 x 256 x 256 \\
   & Conv 3 x 3  & LReLU & 128 x 256 x 256 \\
   & AvgPool     & -     & 128 x 128 x 128   \\
\hline

\multicolumn{4}{|c|}{Model 2 $\downarrow$} \\
\hline

   & Raw RGB images 3& -  &  3 x 128 x 128 \\
   & Concat/$\phi_{\mathit{simple}}$ & -     & 131 x 128 x 128 \\
4. & MinBatchStd & -     & 132 x 128 x 128 \\
   & Conv 3 x 3  & LReLU & 128 x 128 x 128 \\
   & Conv 3 x 3  & LReLU & 256 x 128 x 128 \\
   & AvgPool     & -     & 256 x 64 x 64   \\
\hline

   & Raw RGB images 4& -  &  3 x 64 x 64 \\
   & Concat/$\phi_{\mathit{simple}}$ & -     & 259 x 64 x 64 \\
5. & MinBatchStd & -     & 260 x 64 x 64 \\
   & Conv 3 x 3  & LReLU & 256 x 64 x 64 \\
   & Conv 3 x 3  & LReLU & 512 x 64 x 64 \\
   & AvgPool     & -     & 512 x 32 x 32   \\
\hline

\multicolumn{4}{|c|}{Model 1 $\downarrow$} \\
\hline

   & Raw RGB images 5& -  &  3 x 32 x 32 \\
   & Concat/$\phi_{\mathit{simple}}$ & -     & 515 x 32 x 32 \\
6. & MinBatchStd & -     & 516 x 32 x 32 \\
   & Conv 3 x 3  & LReLU & 512 x 32 x 32 \\
   & Conv 3 x 3  & LReLU & 512 x 32 x 32 \\
   & AvgPool     & -     & 512 x 16 x 16 \\
\hline

   & Raw RGB images 6& -  &  3 x 16 x 16 \\
   & Concat/$\phi_{\mathit{simple}}$ & -     & 515 x 16 x 16 \\
7. & MinBatchStd & -     & 516 x 16 x 16 \\
   & Conv 3 x 3  & LReLU & 512 x 16 x 16 \\
   & Conv 3 x 3  & LReLU & 512 x 16 x 16 \\
   & AvgPool     & -     & 512 x 8 x 8 \\
\hline

   & Raw RGB images 7& -  &  3 x 8 x 8 \\
   & Concat/$\phi_{\mathit{simple}}$ & -     & 515 x 8 x 8 \\
8. & MinBatchStd & -     & 516 x 8 x 8 \\
   & Conv 3 x 3  & LReLU & 512 x 8 x 8 \\
   & Conv 3 x 3  & LReLU & 512 x 8 x 8 \\
   & AvgPool     & -     & 512 x 4 x 4 \\
\hline

   & Raw RGB images 7& -  &  3 x 4 x 4 \\
   & Concat/$\phi_{\mathit{simple}}$ & -     & 515 x 4 x 4 \\
9. & MinBatchStd & -     & 516 x 4 x 4 \\
   & Conv 3 x 3  & LReLU & 512 x 4 x 4 \\
   & Conv 4 x 4  & LReLU & 512 x 1 x 1 \\
   & Fully Connected & Linear & 1 x 1 x 1 \\
\hline

\end{tabular}
\end{center}
\caption{Discriminator Architecture for the MSG-ProGAN and MSG-StyleGAN Models used in training.}
\label{table:dis}
\end{table}


    

\subsection{Architecture Details}
\label{sec:network}

\paragraph{MSG-ProGAN}
Tables \ref{table:gen} and \ref{table:dis} provide the detailed configurations of the generator and the discriminator of MSG-ProGAN respectively.
After every block in the generator, a \texttt{1 x 1 conv} layer is used to convert the output activation volume into an RGB image which is passed onto the discriminator. 
On the discriminator's side, these RGB images are combined with straight path activation volumes using the combine function $\phi$. 
In case of $\phi_{\mathit{simple}}$, a simple channelwise concatenation operation is used (see Table \ref{table:dis}). 
For the $\phi_{\mathit{lin\_cat}}$ variant of the combine function, a \texttt{1 x 1} conv layer is used to project the RGB images into activation space which is then followed by channelwise concatenation operation. 
The number of channels output by the \texttt{1 x 1} conv layer is equal to half of the output channels in that block of the discriminator, e.g. for block 3 (see Table \ref{table:dis}), the output of the \texttt{1 x 1} conv layer is \texttt{32 x 256 x 256} and the output of $\phi_{\mathit{lin\_cat}}$ operation is \texttt{96 x 256 x 256} (32 + 64). 
Finally, for the $\phi_{\mathit{cat\_lin}}$, the RGB images are first concatenated with the straight path activation volumes followed by a \texttt{1 x 1} conv layer. The number of channels output by this \texttt{1 x 1} conv layer is again equal to the prevalent number of channels in that block (e. g. 64 for block 3).

\emph{Model 1}, \emph{Model 2} and \emph{Model 3} blocks of the generator (Tab~\ref{table:gen}) are used to synthesize \texttt{32 x 32}, \texttt{128 x 128} and \texttt{256 x 256} sized images respectively. And, after every \texttt{3 x 3 conv} operation the feature vectors are normalized according to the PixNorm \cite{proGAN2018} scheme (only for the generator).

\paragraph{MSG-StyleGAN}
The MSG-StyleGAN model uses all the modifications proposed by StyleGAN \cite{styleGAN2018} to the ProGANs \cite{proGAN2018} architecture except the mixing regularization. 
Similar to MSG-ProGAN, we use a \texttt{1 x 1} conv layer to obtain the RGB images output from every block of the StyleGAN generator leaving everything else (mapping network, non-traditional input and style adaIN) untouched.
The discriminator architecture is same as the ProGANs (and consequently MSG-ProGAN, Tab.~\ref{table:dis}) discriminator.

\subsection{Additional Qualitative Results}
\begin{figure*}[p]
\begin{center}
\includegraphics[width=0.8\linewidth]{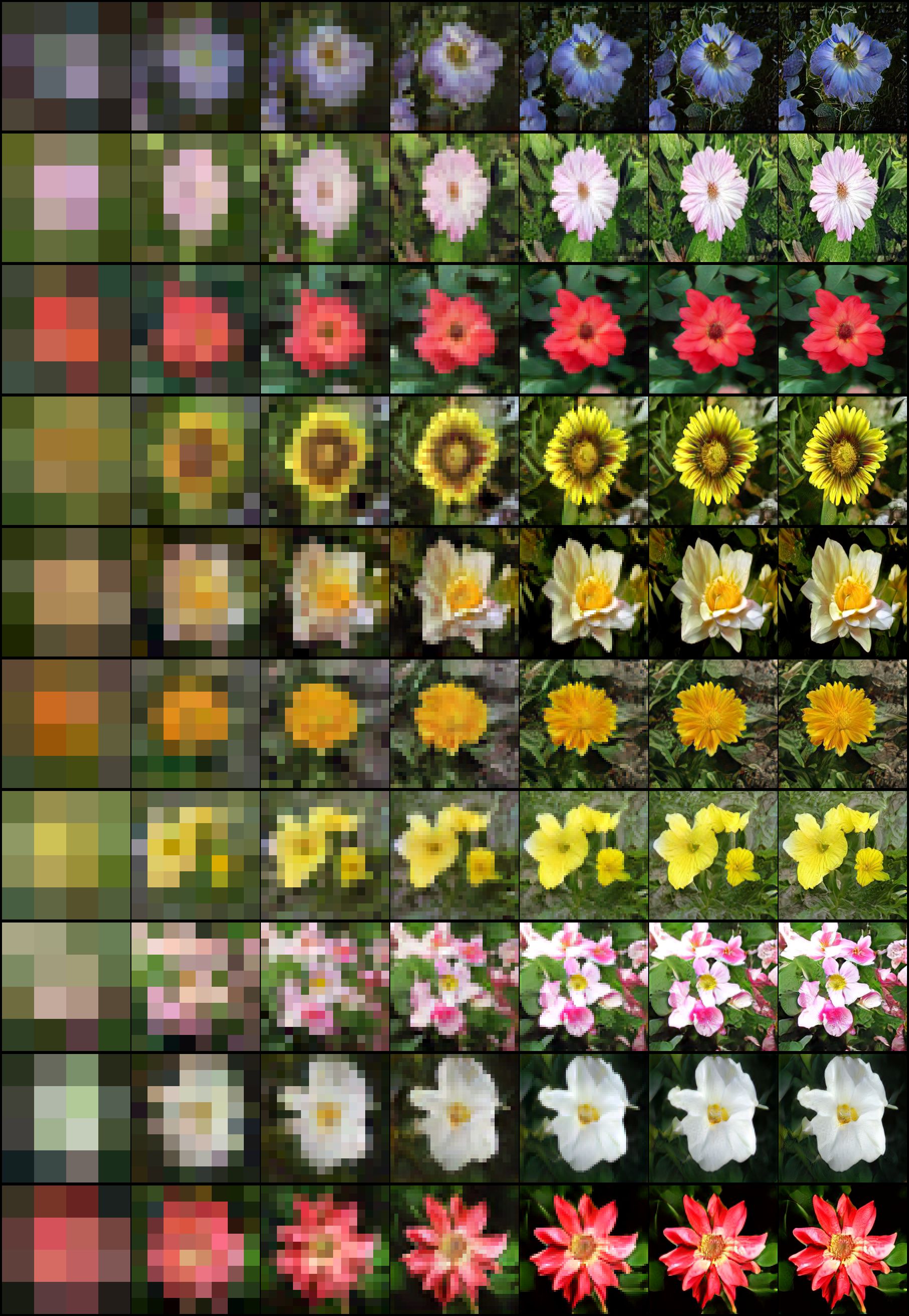}
\end{center}
   \caption{Random samples generated at all $7$ resolutions for the Oxford102 flowers dataset.}
\label{fig:flowers_all_res}
\end{figure*}

\begin{figure*}[p]
\begin{center}
\includegraphics[width=0.9\linewidth]{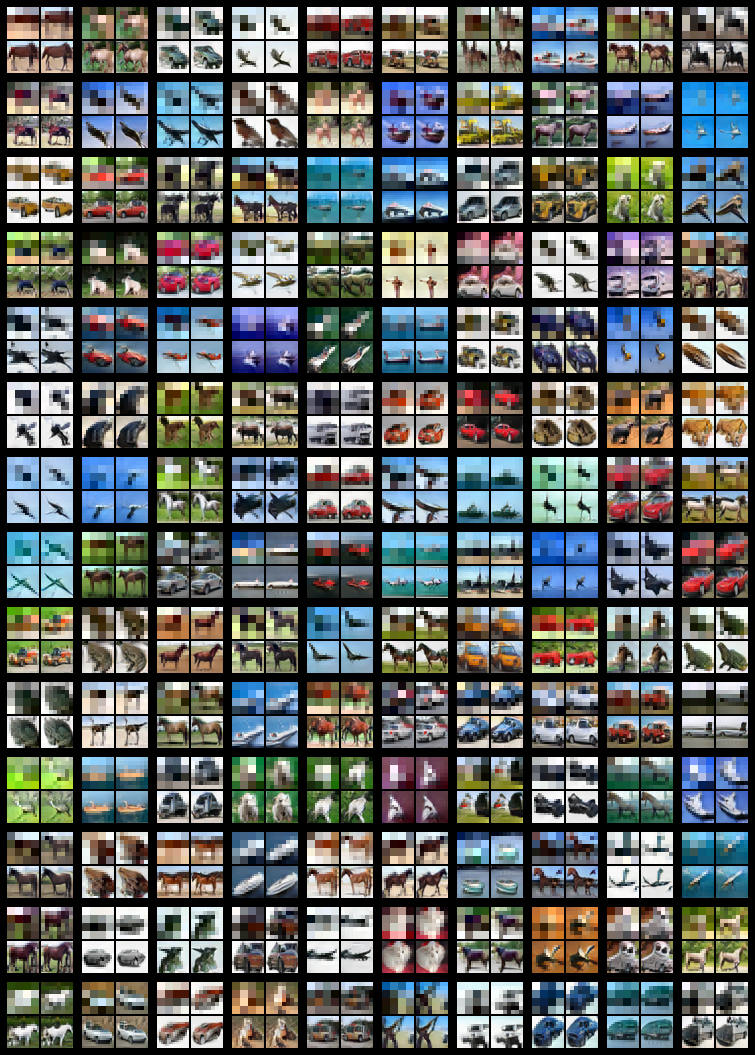}
\end{center}
   \caption{Random samples generated at all $4$ resolutions for the CIFAR-10 dataset.}
\label{fig:cifar_all_res}
\end{figure*}

\begin{figure*}[p]
\begin{center}
\includegraphics[width=0.9\linewidth]{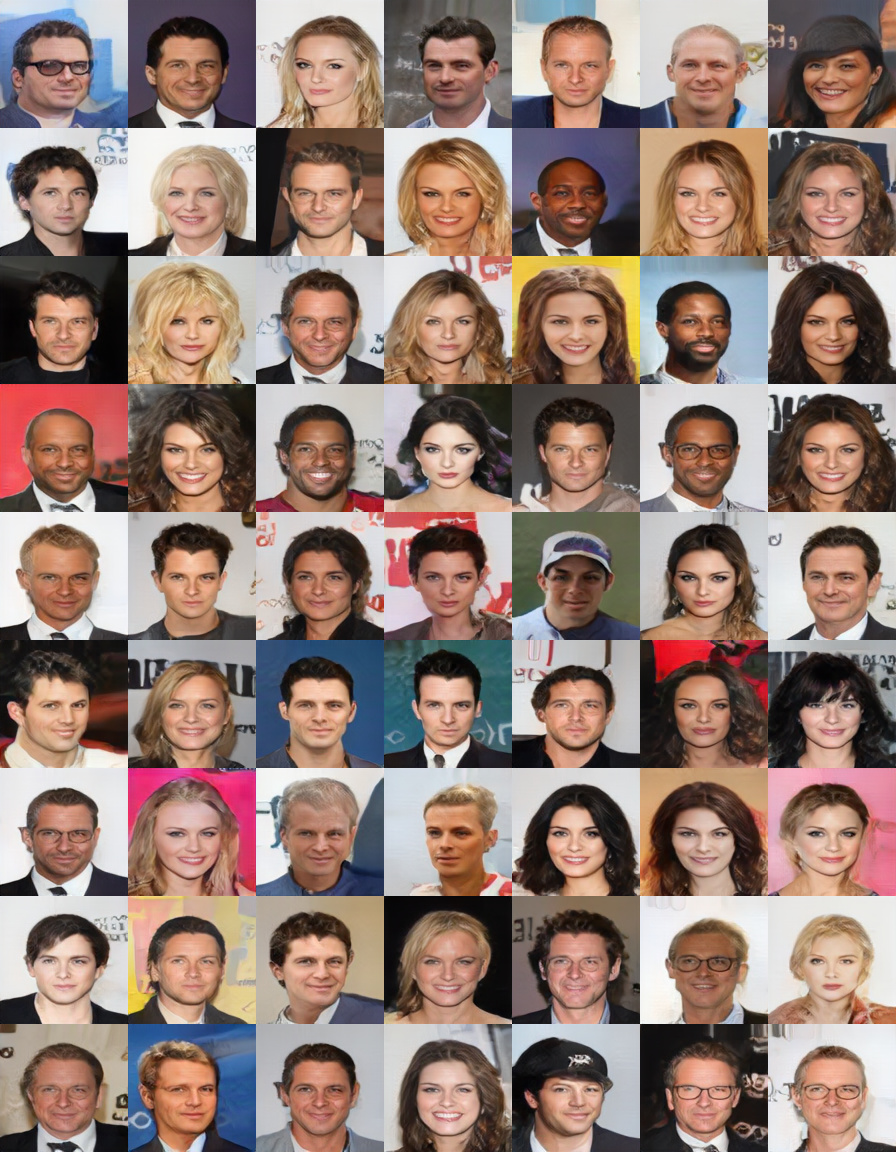}
\end{center}
   \caption{Random generated CelebA Faces at resolution \texttt{128 x 128}.}
\label{fig:celeba_128}
\end{figure*}

\begin{figure*}[p]
\begin{center}
\includegraphics[width=0.9\linewidth]{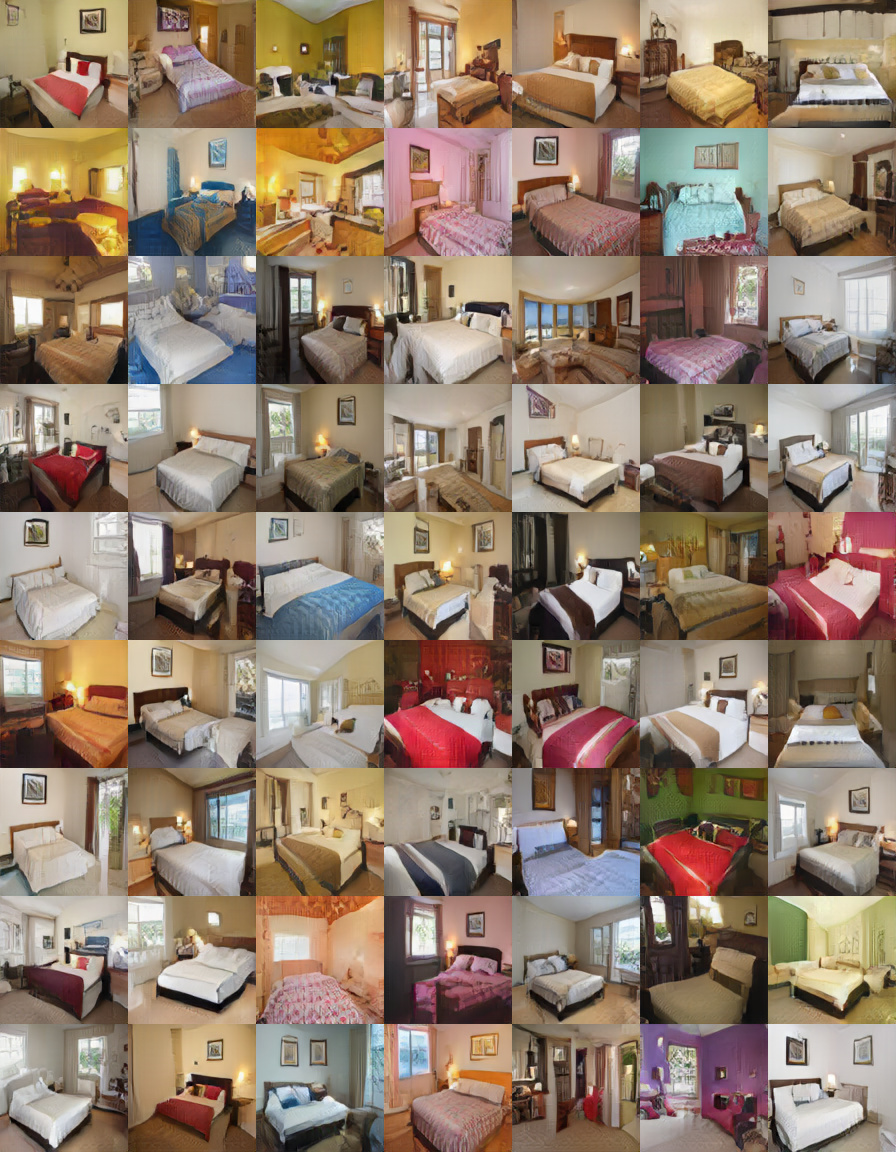}
\end{center}
   \caption{Random generated LSUN bedrooms at resolution \texttt{128 x 128}.}
\label{fig:lsun_128}
\end{figure*}

\begin{figure*}[p]
\begin{center}
\includegraphics[width=0.9\linewidth]{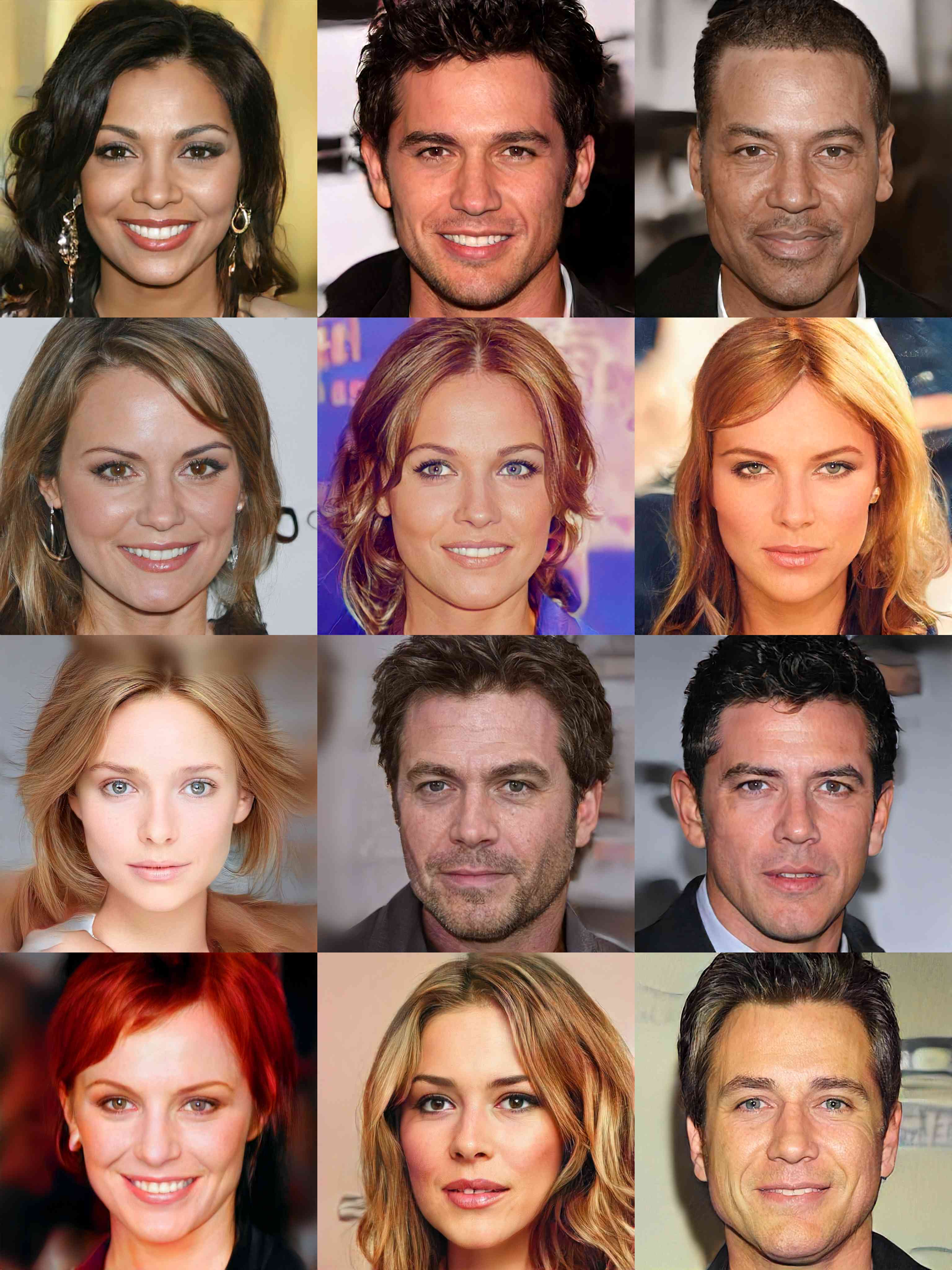}
\end{center}
  \caption{Random generated CelebA-HQ Faces at resolution \texttt{1024 x 1024}.}
\label{fig:celebahq}
\end{figure*}

Here we include additional results for further empirical validation. 
We show full resolution results from MSG-StyleGAN for the \texttt{256 x 256} Oxford102 flower dataset, and the MSG-ProGAN architecture for the \texttt{128 x 128} CelebA and LSUN bedroom datasets. 
The CelebA model was trained for $28M$ real images and obtained an FID of \textbf{$8.86$}. 
Because of the huge size of the LSUN bedrooms dataset (30M), we trained it for $150M$ real images (roughly 5 epochs) which resulted in an FID of $18.32$. 
Figures \ref{fig:celeba_128} and \ref{fig:lsun_128} show the \texttt{128 x 128} (highest resolution) samples generated for the CelebA and LSUN bedrooms datasets respectively. 
Figure \ref{fig:flowers_all_res} and Fig \ref{fig:cifar_all_res} shows samples generated by the MSG-StyleGAN model at all resolutions on the Oxford Flowers and Cifar-10 datasets respectively. Figure \ref{fig:celebahq} shows additional qualitative results (random samples) from the CelebA-HQ dataset, trained using our \emph{Model full} architecture at \texttt{1024 x 1024} resolution.

\begin{figure*}[t]
    \begin{subfigure}[t]{.5\textwidth}
        \centering
        \includegraphics[width=1.0\linewidth]{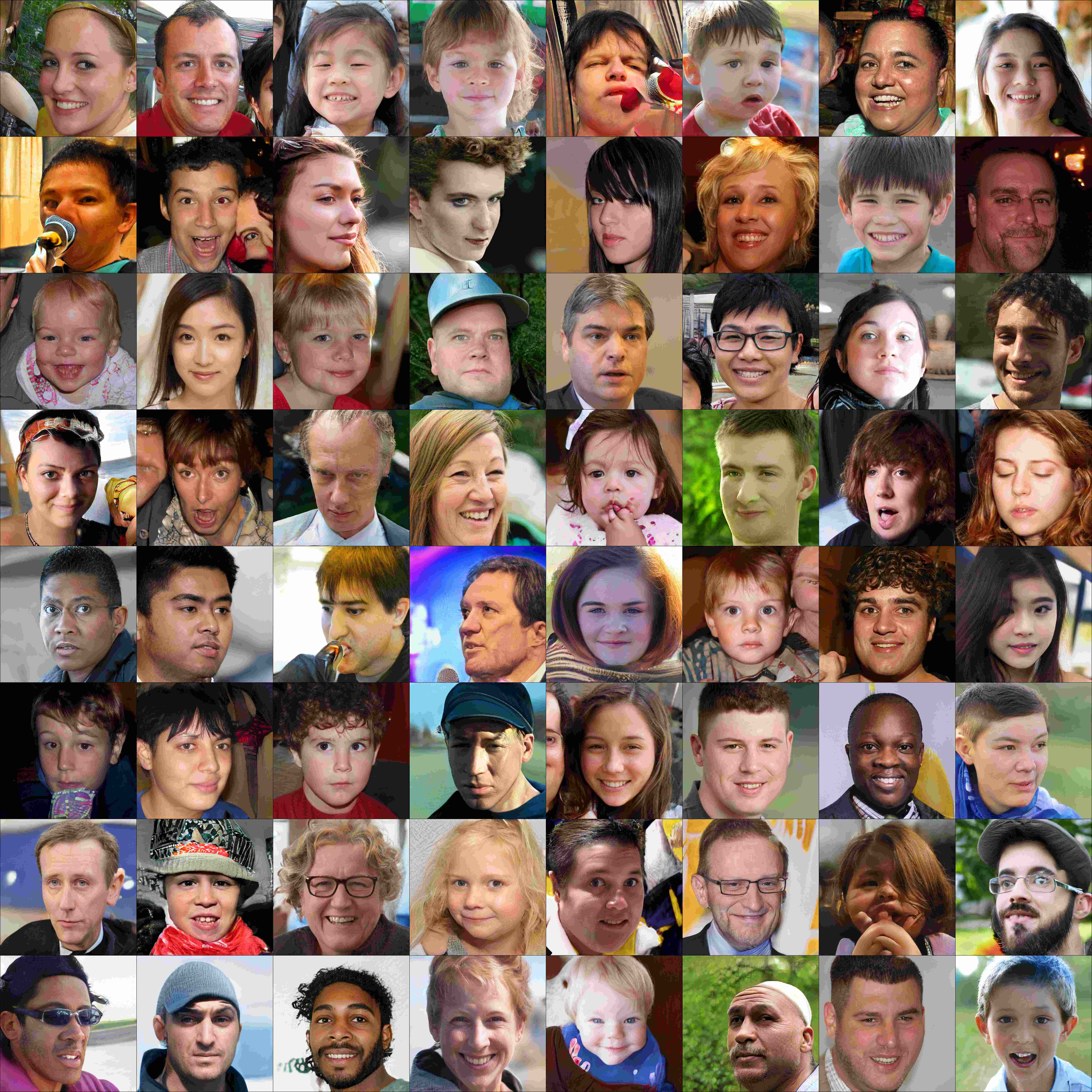}
        \caption{StyleGAN generated images}
    \end{subfigure}
    \hfill
    \begin{subfigure}[t]{.5\textwidth}
        \centering
        \includegraphics[width=1.0\linewidth]{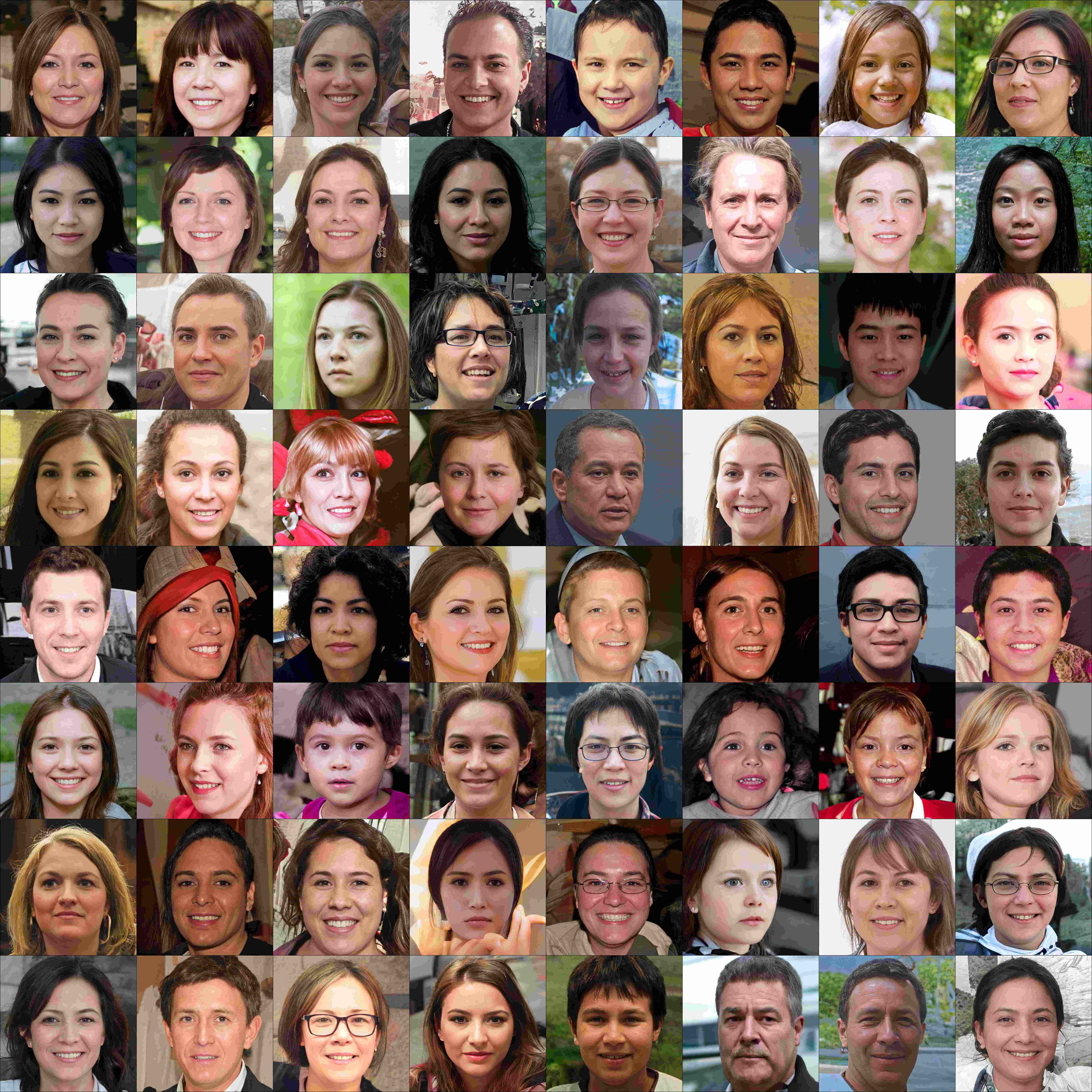}
        \caption{MSG-StyleGAN generated images}
    \end{subfigure}
    
    \caption{
    Random generated samples for qualitative comparison between StyleGAN \cite{styleGAN2018} and MSG-StyleGAN. 
    All the samples were generated \textbf{\emph{without}} truncating the input latent space for both because the FID calculation is done on non-truncated latent spaces. Best viewed zoomed in.
    }
    \label{fig:qualitative_comparison}
\end{figure*}

\subsection{Observations}
In this section, we present some of our observations and hypotheses about the differences in results generated by our method and StyleGAN. 
We show an overview of randomly selected samples from both models in Fig \ref{fig:qualitative_comparison}.
In our analysis of the results, we find that while the actual resulting image quality is very close, StyleGAN samples exhibit \emph{slightly} higher variation in terms of pose. 
In contrast, MSG-StyleGAN results are \emph{slightly} more globally consistent and more realistic.
This trade-off between diversity and result quality is widely reported~\cite{zhu2017toward}, and may explain some of the difference in FID score. 
Further investigation into methods to control either axis (realism vs diversity), and the impact this has on the FID score, would be an interesting avenue for future work. 

\begin{figure*}[t]
\centering
\def\figwidth{0.19\linewidth}
\begin{tabular}{*{5}{c@{\hspace{.2mm}}}}
\includegraphics[width=\figwidth]{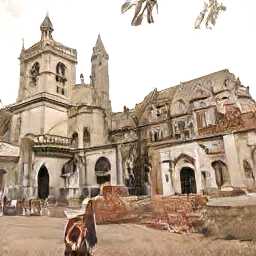}& 
\includegraphics[width=\figwidth]{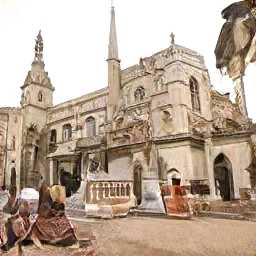}&
\includegraphics[width=\figwidth]{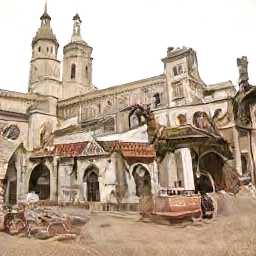}&
\includegraphics[width=\figwidth]{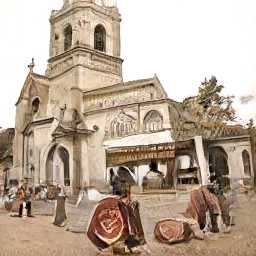}&
\includegraphics[width=\figwidth]{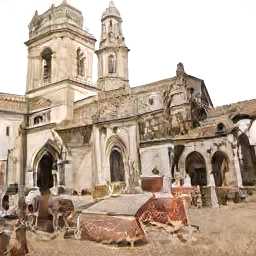}\\
\includegraphics[width=\figwidth]{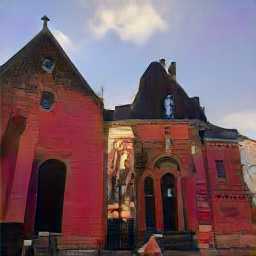}& 
\includegraphics[width=\figwidth]{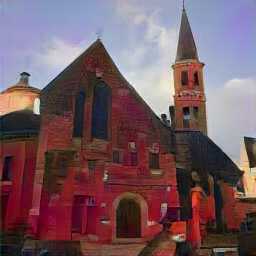}&
\includegraphics[width=\figwidth]{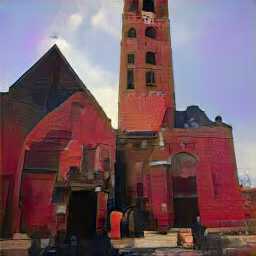}&
\includegraphics[width=\figwidth]{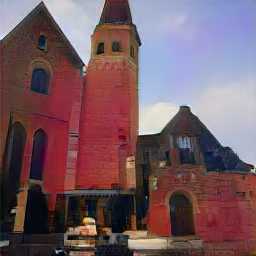}&
\includegraphics[width=\figwidth]{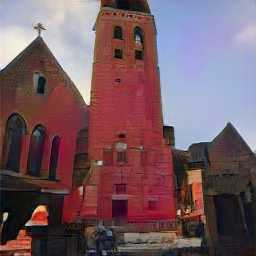}\\
\end{tabular}
\caption{LSUN Church images generated by StyleGAN (top) and MSG-StyleGAN (bottom) using different realizations of the per-pixel noise while keeping the input latent vectors constant.}
\label{fig:stylegan_churches}
\label{fig:msg_stylegan_churches}
\end{figure*}

We also conducted experiments investigating the role that the pixelwise noise added to each block of the StyleGAN generator plays in image generation. 
We found that on non-face datasets, these noise layers model \emph{semantic} aspects of the images and not just stochastic variations, as was their initial intent~\cite{styleGAN2018} (see Fig \ref{fig:stylegan_churches}).
We observed that MSG-StyleGAN also shows this type of effect, although to a slightly less degree. 
We conjecture that this disentanglement between the stochastic and semantic features is more straightforward for the face modelling task (e.g., on CelebA-HQ and FFHQ datasets), and the different models sensitivity to this noise could contribute to some of the the performance differences we observe as well, on face vs non-face datasets. 

\begin{figure*}[t]
\includegraphics[width=1.0\linewidth]{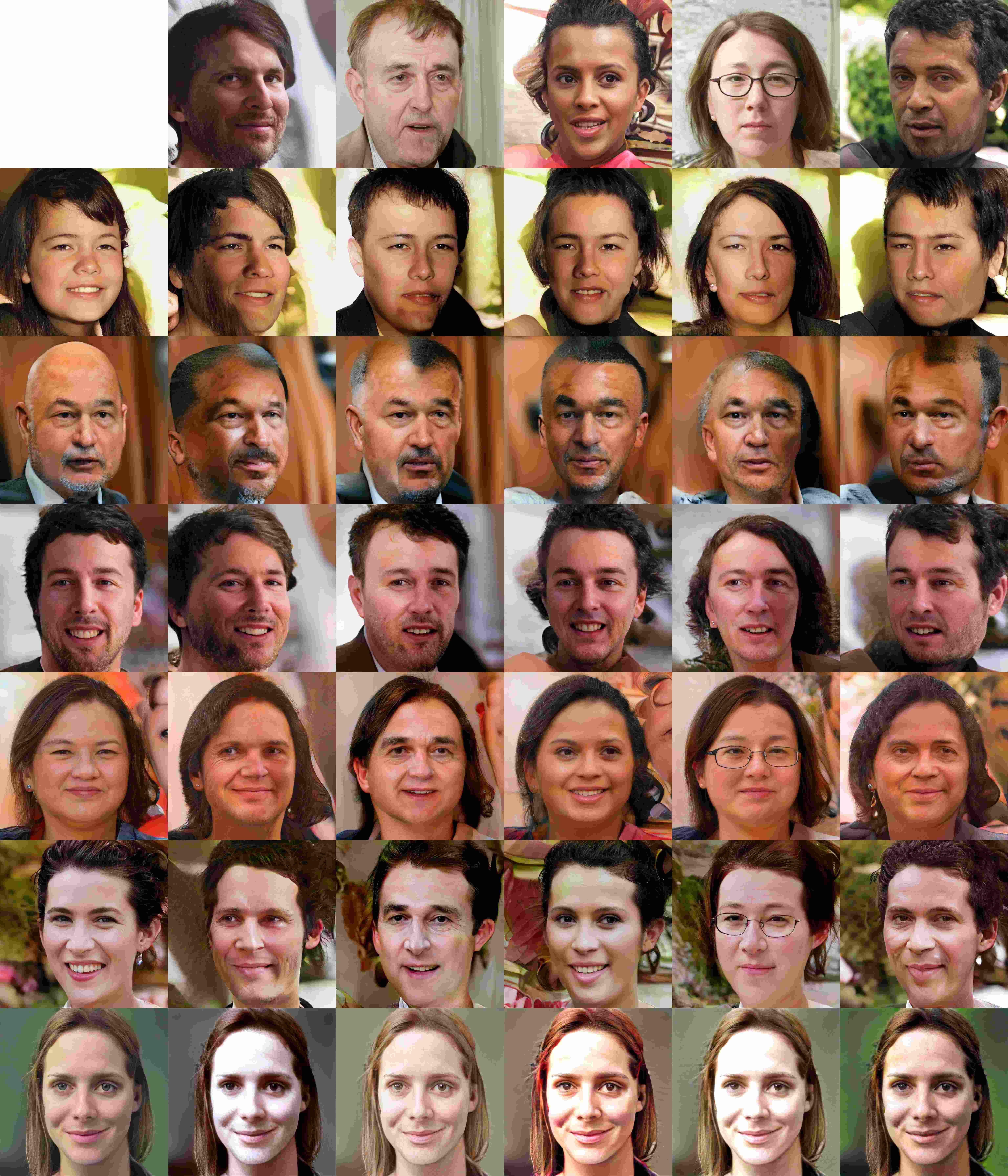}
\caption{Images generated by mixing the styles coming from two different latent vectors at different levels (granularity) of generation. As in StyleGAN~\cite{styleGAN2018}, the first column images are source 1 and first row are source 2. Rows numbered 2, 3, and 4 have the mixing at resolutions (\texttt{4 x 4} and \texttt{8 x 8}), while rows 5 and 6 at (\texttt{16 x 16} and \text{32 x 32}), and the row 6 images are generated by swapping the source 2 latents at resolutions (\texttt{64 x 64} till \texttt{1024 x 1024}).}
\label{fig:mixing_figure}
\end{figure*}

As mentioned in the discussion section of the main paper, we do not use the mixing regularization technique described in the StyleGAN \cite{styleGAN2018} work (the question of how to integrate such a regularization is an interesting direction for future work).
However, we note that in spite of not using it, the model still learns to disentangle high level semantic features of the images due to the scale based constraint (see Fig. \ref{fig:mixing_figure}). 
As apparent from the figure, the high level mixing is much more coherent and generates more visually realistic results; while lower level mixing often generates incorrect visual cues, such as improper lighting and unbalanced hair. 
This shows that performance gains might be possible by ensuring proper style-based mixing at the low (coarse-grained) level of generation.

